\documentclass{article}

\usepackage[final,nonatbib]{neurips_2020}

\usepackage[utf8]{inputenc} 
\usepackage[T1]{fontenc}    
\usepackage{booktabs} 
\usepackage{times}  
\usepackage{helvet}  
\usepackage{courier}  

\usepackage{graphicx}
\usepackage{amsmath}
\usepackage{amsthm}
\usepackage{amssymb}
\usepackage{algorithm}
\usepackage{algorithmic}
\usepackage[tight,footnotesize]{subfigure}
\usepackage{url}
\usepackage{bm}
\usepackage{nicefrac}
\usepackage{subfigure}
\usepackage{caption}
\usepackage{enumitem}
\usepackage{multirow}
\usepackage{color}
\usepackage{titlesec}
\usepackage{wrapfig}
\usepackage{caption}

\newtheorem{definition}{Definition}

\usepackage{array}
\newcolumntype{C}[1]{>{\centering\let\newline  \\\arraybackslash\hspace{0pt}}m{#1}}

\title{Interstellar: Searching Recurrent Architecture for 
	Knowledge Graph Embedding}

\author{%
	Yongqi Zhang$^{1,3}$
	\quad
	Quanming Yao$^{1,2}$
	\quad
	Lei Chen$^3$\\
	$^1$4Paradigm Inc. \\
	$^2$Department of Electronic Engineering, Tsinghua University \\
	$^3$Department of Computer Science and Engineering, HKUST \\
	\{zhangyongqi,yaoquanming\}@4paradigm.com, leichen@cse.ust.hk
}

\begin{document}

\maketitle

\begin{abstract}		
Knowledge graph (KG) embedding is well-known in learning representations of KGs.
Many models have been proposed to learn the interactions between entities and relations
of the triplets.
However, long-term information among multiple triplets
is also important to KG.
In this work,
based on the relational paths, which are composed of a sequence of triplets,
we define 
the Interstellar as a recurrent neural architecture search problem for the short-term and long-term information 
along the paths.
First,
we analyze the difficulty of using a unified model to work as the Interstellar.
Then,
we propose to search for recurrent architecture as the Interstellar for different KG tasks.
A case study on synthetic data illustrates the importance of the defined search problem.
Experiments on real datasets demonstrate the effectiveness of the searched models
and the efficiency of the proposed hybrid-search algorithm.
\footnote{Code is available at \url{https://github.com/AutoML-4Paradigm/Interstellar},
	and correspondence is to Q. Yao.
	\vspace{-10px}
}
\end{abstract}

\section{Introduction}

\begin{figure}[t]
\centering
\vspace{-8px}
\subfigure[An example of Knowledge Graph.]
{\includegraphics[width=0.48\columnwidth]{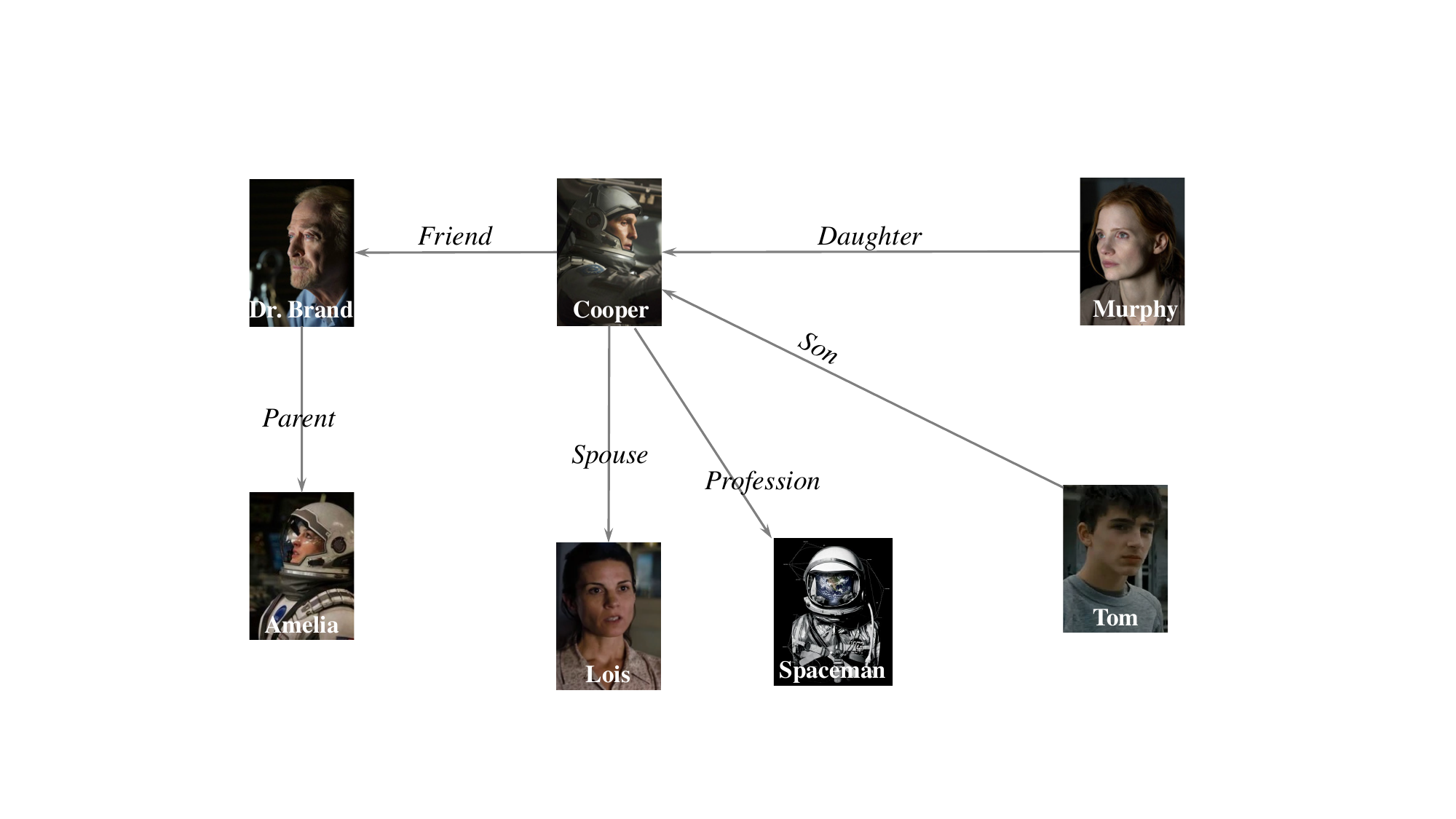}
	\label{fig:eg:kg}}
\hfill
\subfigure[Examples of relational paths.]
{\includegraphics[width=0.47\columnwidth]{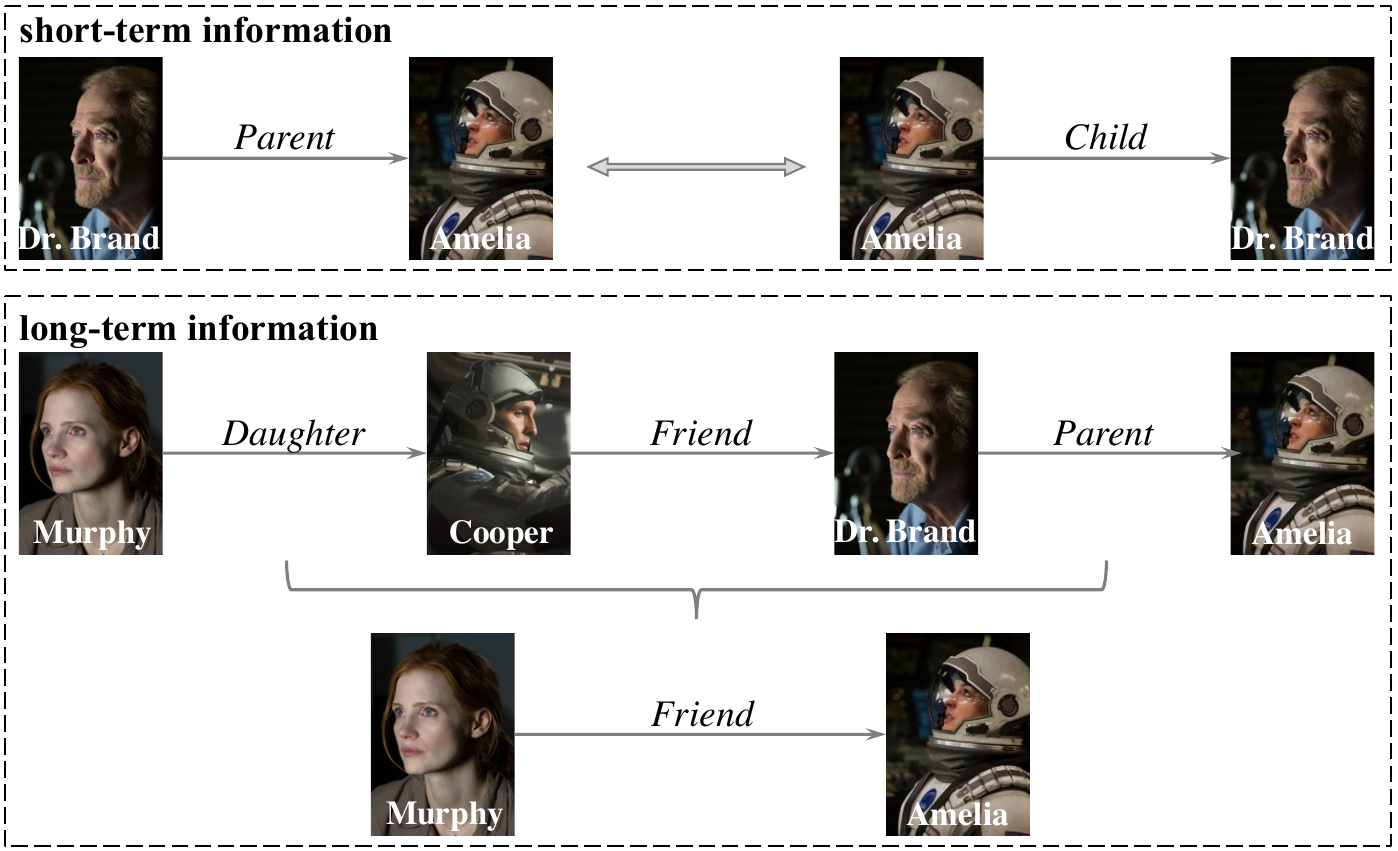}
	\label{fig:eg:paths}}
\vspace{-10px}
\caption{
	Short-term information is represented by a single triplet.
	Long-term information passes across multiple triplets.
	The two kinds of information in KGs can be preserved in the relational path.}
\label{fig:kg}
\vspace{-15px}
\end{figure}

Knowledge Graph (KG) \cite{auer2007dbpedia,socher2013reasoning,wang2017knowledge} is a special kind of graph with many relational facts.
It has inspired many knowledge-driven applications,
such as 
question answering \cite{lukovnikov2017neural,ren2019query2box},
medical diagnosis \cite{zhang2018generative},
and recommendation \cite{liu2019survey}.
An example of the KG is in Figure~\ref{fig:eg:kg}.
Each relational fact in KG
is represented as a triplet in the form of \textit{(subject entity, relation, object entity)},
abbreviated as $(s, r, o)$.
To learn from the KGs and benefit the downstream tasks,
embedding based methods, 
which learn low-dimensional vector representations of the entities and relations,
have recently developed as a promising direction to serve this purpose \cite{bordes2013translating,guo2019learning,sun2018bootstrapping,wang2017knowledge}.

Many efforts have been made on modeling the plausibility of triplets $(s,r,o)$s through learning embeddings.
Representative works are triplet-based models,
such as
TransE \cite{bordes2013translating},
ComplEx \cite{trouillon2017knowledge}, 
ConvE \cite{dettmers2017convolutional},
RotatE \cite{sun2019rotate},
AutoSF \cite{zhang2020autosf},
which
define different embedding spaces and learn on single triplet $(s,r,o)$.
Even though these models
perform well in capturing short-term semantic information inside the triplets in KG,
they still cannot capture the information among multiple triplets.

In order to better capture the complex information in KGs,
the relational path is introduced as a promising format to 
learn composition of relations \cite{guu2015traversing,lin2015modeling,sadeghian2019drum}
and long-term dependency of triplets \cite{lao2011random,das2016chains,guo2019learning,wang2020entity}.
As in Figure~\ref{fig:eg:paths},
a relational path is defined as a set of $L$ triplets 
$(s_1, r_1, o_1), (s_2, r_2, o_2),\dots,(s_L, r_L, o_L)$,
which are connected head-to-tail in sequence, i.e. $o_i =s_{i+1}, \forall i=1\dots L-1$.
The paths not only preserve every single triplet but also can capture the dependency among a sequence of triplets.
Based on the relational paths,
the triplet-based models can be compatible by
working on each triplet $(s_i, r_i, o_i)$ separately.
TransE-Comp \cite{guu2015traversing}
and PTransE \cite{lin2015modeling}
learn the composition relations on the relational paths.
To capture the long-term information in KGs,
Chains \cite{das2016chains} and RSN \cite{guo2019learning}
design customized RNN to leverage all the entities and relations along path.
However, the RNN models still overlook the semantics inside each triplet \cite{guo2019learning}.
Another type of models leverage Graph Convolution Network (GCN) \cite{kipf2016semi} to extract structural information in KGs,
e.g. R-GCN, GCN-Align~\cite{wang2018cross}, CompGCN \cite{vashishth2019composition}.
However, GCN-based methods do not scale well since the entire KG needs to be processed
and it has large sample complexity \cite{garg2020generalization}.

In this paper, we observe that the relational path is an important and effective data structure 
that can preserve both short-term and long-term information in KG.
Since the semantic patterns and the graph structures in KGs are diverse \cite{wang2017knowledge},
how to leverage the short-term and long-term information for a specific KG task is non-trivial.
Inspired by the success of neural architecture search (NAS) \cite{elsken2019neural},
we propose to search recurrent architectures as the \textit{Interstellar} to learn from the relational path.
The contributions of our work are summarized as follows:

\begin{enumerate}[parsep=0pt,partopsep=0pt,leftmargin=15px]
	\item 
	We analyze the difficulty and importance of using the relational path
	to learn the short-term and long-term information in KGs.
	Based on the analysis, 
	we define the Interstellar as a recurrent network 
	to process the information along the relational path.
	
	\item
	We formulate the above problem as a NAS problem and
	propose a domain-specific search space.
	Different from searching RNN cells, the recurrent network in our space is specifically designed for KG tasks
	and covers many human-designed embedding models.
	
	\item 
	We identify the problems of adopting stand-alone and one-shot search algorithms for our search space.
	This motivates us to design a hybrid-search algorithm to search efficiently.
	
	\item 
	We use a case study on the synthetic data set to show the reasonableness of our search space.
	Empirical experiments on entity alignment and link prediction tasks 
	demonstrate the effectiveness of the searched models and the efficiency of the search algorithm.
\end{enumerate}

\textbf{Notations.}
We denote vectors by lowercase boldface, and matrix by uppercase boldface.
A KG $\mathcal G= \left(\mathcal E, \mathcal R, \mathcal S\right)$
is defined by the set of entities $\mathcal E$,
relations $\mathcal R$ and triplets $\mathcal S$.
A triplet $(s, r, o)\in\mathcal S$
represents a relation $r$ that links from the subject entity $s$ to the object entity $o$.
The embeddings in this paper are denoted as boldface letters of indexes,
e.g. $\mathbf s, \mathbf r, \mathbf o$ are embeddings of $s, r, o$.
``$\odot$" is the element-wise multiply
and ``$\otimes$" is the Hermitian product \cite{trouillon2017knowledge} in complex space.

\section{Related Works}

\subsection{Representation Learning in Knowledge Graph (KG)}
\label{ssec:rel:kg}

Given a single triplet $(s, r, o)$,
TransE \cite{bordes2013translating}
models the relation $r$ as a translation vector from subject entity $s$ to object entity $o$,
i.e.,
the embeddings satisfy $\mathbf s+\mathbf r\approx\mathbf o$.
The following works
DistMult \cite{yang2014embedding},
ComplEx \cite{trouillon2017knowledge},
ConvE \cite{dettmers2017convolutional},
RotatE \cite{sun2019rotate},
etc.,
interpret the interactions among embeddings $\mathbf s, \mathbf r$ and $\mathbf o$ in different ways.
All of them learn embeddings based on single triplet.

In KGs,
a relational path is a sequence of triplets.
PTransE \cite{lin2015modeling} and TransE-Comp \cite{guu2015traversing} propose to learn 
the composition of relations $(r_1, r_2, \dots, r_n)$.
In order to combine more pieces of information in KG,
Chains \cite{das2016chains} and RSN \cite{guo2019learning} are proposed to jointly learn 
the entities and relations along the relational path.
With different connections and combinators,
these models process short-term and long-term information in different ways.

Graph convolutional network (GCN) \cite{kipf2016semi} have recently been developed as a promising method to learn from graph data.
As a special instance of graph,
GCN has also been introduced in KG learning,
e.g.,
R-GCN \cite{schlichtkrull2018modeling}, GCN-Align \cite{wang2018cross}, VR-GCN \cite{ye2019vectorized}
and CompGCN \cite{vashishth2019composition}.
However, these models are hard to scale well since the whole KG should be loaded.
 for each training iteration.
Besides,
GCN has been theoretically proved to have worse generalization guarantee than RNN 
in sequence learning tasks~(Section 5.2 in \cite{garg2020generalization}).
Table~\ref{tab:human-design} summarizes
above works and compares them with the proposed Interstellar,
which is a NAS method customized to path-based KG representation learning.

\begin{table}[t]
	\centering
	\vspace{-8px}
	\caption{The recurrent function of existing KG embedding models.
		We represent the triplet/path-based models by Definition~\ref{def:path}.
		$\mathcal N(\cdot)$ denotes the  neighbors of an entity.
		$\mathbf W\in\mathbb R^{d\times d}$'s are different weight matrices.
		$\sigma$ is a non-linear activation function.
		``cell'' means a RNN cell \cite{dasgupta1996sample}, like GRU \cite{chung2015gated} LSTM \cite{sundermeyer2012lstm} etc.
		For mini-batch complexity, $m$ is the batch size, $d$ is the embedding dimension.}
	\setlength\tabcolsep{5px}
	\label{tab:human-design}
	\small
	\renewcommand{\arraystretch}{1.1}
	\begin{tabular}{c|  c | c | c | c}
		\hline
		type & \multicolumn{2}{c|}{model}                          &       unit function                                    &   complexity                        \\ \hline
		\multirow{2}{*}{triplet-based} &   \multicolumn{2}{c|}{TransE \cite{bordes2013translating}}          & $\mathbf v_t= \mathbf s_t+\mathbf r_t, \mathbf h_t = 0$                        &       $O(md)$             \\ \cline{2-5}
		& \multicolumn{2}{c|}{ComplEx \cite{trouillon2017knowledge}}          & $\mathbf v_t= \mathbf s_t\otimes \mathbf r_t, , \mathbf h_t = 0$                                                                                  &       $O(md)$              \\ \hline
		\multirow{2}{*}{GCN-based} & 	\multicolumn{2}{c|}{R-GCN~\cite{schlichtkrull2018modeling}} &   
		$\mathbf s_t = \sigma (\mathbf s_{t-1}+\sum_{s'\in\mathcal N(s)}\mathbf W^{(r)}_t\mathbf s'_{t-1})$  &   $O(|\mathcal E||\mathcal R|d)$ \\ \cline{2-5}
		& \multicolumn{2}{c|}{GCN-Align~\cite{wang2018cross}} &  
		$\mathbf s_t = \sigma (\mathbf s_{t-1}+\sum_{s'\in\mathcal N(s)}\mathbf W_t\mathbf s'_{t-1})$  &   $O(|\mathcal E|d)$ \\ \hline 
		\multirow{6}{*}{path-based} & 	 & add     & 
		$\mathbf v_t= \mathbf h_t, \mathbf h_t = \mathbf h_{t-1}+\mathbf r_t$                  &        $O(mLd)$             \\ \cline{3-5}
		&	PTransE \cite{lin2015modeling}& multiply   & $\mathbf v_t= \mathbf h_t, \mathbf h_t = \mathbf h_{t-1} \odot \mathbf r_t$                                 &        $O(mLd)$          \\  \cline{3-5}
		&	& RNN      & $\mathbf v_t  =  \mathbf h_t, \mathbf h_t \! = \! \text{cell}\! \left(\mathbf r_t, \mathbf h_{t-1}\right )$                      &    $O(mLd^2)$          \\   \cline{2-5}
		& \multicolumn{2}{c|}{Chains \cite{das2016chains}}              & $\mathbf v_t = \mathbf h_t, \mathbf h_t \! = \! \text{cell}\!\left(\mathbf s_t,\mathbf r_t, \mathbf h_{t-1}\right)$             &     $O(mLd^2)$     \\  \cline{2-5}
		&  \multicolumn{2}{c|}{RSN \cite{guo2019learning}}         & $\!\mathbf v_t \!\!=\!\! \mathbf W_1\mathbf s_t \!\!+\!\! \mathbf W_2\mathbf h_t, \mathbf h_{t} \! = \!
		\text{cell}\!\left(\mathbf r_t, \text{cell}(\mathbf s_t, \mathbf h_{t-1})\right)$  & $O(mLd^2)$  \\ \cline{2-5}
		&  \multicolumn{2}{c|}{Interstellar}                        &                                               {a searched recurrent network}                  &           $O(mLd^2)$               \\ \hline
	\end{tabular}
	\vspace{-10px}
\end{table}

\subsection{Neural Architecture Search (NAS)}
\label{ssec:rel:nas}
Searching for better neural networks by NAS techniques 
have broken through the bottleneck in manual architecture designing~\cite{elsken2019neural,automl_book,sciuto2019evaluating,zoph2017neural,yao2018taking}.
To guarantee  effectiveness and efficiency,
the first thing we should care about is the search space.
It defines what architectures can be represented in principle, 
like CNN or RNN.
In general, the search space should be powerful but also tractable.
The space of CNN has been developed 
from searching the macro architecture \cite{zoph2017neural},
to micro cells \cite{liu2018darts}
and further to the larger and sophisticated cells \cite{tan2019efficientnet}.
Many promising architectures have been searched 
to outperform human-designed CNNs in literature \cite{akimoto2019adaptive,xie2018snas}.
However, designing the search space for recurrent neural network attracts little attention.
The searched architectures mainly focus on cells rather than connections among cells \cite{pham2018efficient,zoph2017neural}.

To search efficiently,
evaluation method, which provides feedback signals,
and search algorithm, which guides the optimization direction,
should be simultaneously considered.
There are two important approaches for NAS.
1) Stand-alone methods,
i.e. separately training and evaluating each model from scratch,
are the most guaranteed way to compare different architectures,
whereas very slow.
2) One-shot search methods,
e.g., DARTS~\cite{liu2018darts}, 
ASNG~\cite{akimoto2019adaptive} and NASP~\cite{yao2020efficient},
have recently become the most popular approach that can efficiently find good architectures.
Different candidates are approximately evaluated in a supernet with parameter-sharing (PS).
However, PS is not always reliable,
especially in complex search spaces
like the macro space of CNNs and RNNs \cite{bender2018understanding,pham2018efficient,sciuto2019evaluating}.
\vspace{-3px}

\section{The Proposed Method}
\label{sec:space}

As in Section~\ref{ssec:rel:kg},
the relational path is informative in representing knowledge in KGs.
Since the path is a sequence of triplets with varied length,
it is intuitive to use Recurrent Neural Network (RNN),
which is known to have a universal approximation ability~\cite{schafer2006recurrent}, 
to model the path as language models \cite{sundermeyer2012lstm}.
While RNN can capture the long-term information along steps \cite{mikolov2010recurrent,chung2015gated},
it will overlook domain-specific properties like the semantics inside each triplet without a customized architecture~\cite{guo2019learning}.
Besides,
what kind of information should we leverage varies from tasks~\cite{wang2017knowledge}.
Finally,
as proved in \cite{garg2020generalization},
RNN has better generalization guarantee than GCN.
Thus,
to design a proper KG model,
we define the path modeling as a NAS problem for RNN here.

\subsection{Designing a Recurrent Search Space}
\label{ssec:Interstellar}

To start with,
we firstly define a general recurrent function (Interstellar) on the relational path.

\begin{definition}[Interstellar]
\label{def:path}
An Interstellar processes the embeddings of $\mathbf s_1, \mathbf r_1$ to $\mathbf s_L, \mathbf r_L$ recurrently.
In each recurrent step $t$, the Interstellar combines embeddings of $\mathbf s_t$, $\mathbf r_t$
and the preceding information $\mathbf h_{t-1}$
to get an output $\mathbf v_t$.
The Interstellar is formulated as a recurrent function
\begin{equation}
\left[ \mathbf{v}_t, \mathbf{h}_t \right] 
= f(\mathbf s_t, \mathbf r_t, \mathbf h_{t-1}),
\quad
\forall t = 1 \dots L,
\label{eq:recurrent}
\end{equation}
where $\mathbf h_t$ is the recurrent hidden state
and $\mathbf h_0=\mathbf s_1$.
The output $\mathbf v_t$ is to predict object entity $o_t$.
\end{definition}

\begin{figure}[ht]
	\vspace{-7px}
	\begin{minipage}{0.50\textwidth}
	\centering
	\includegraphics[width = 0.95\textwidth]{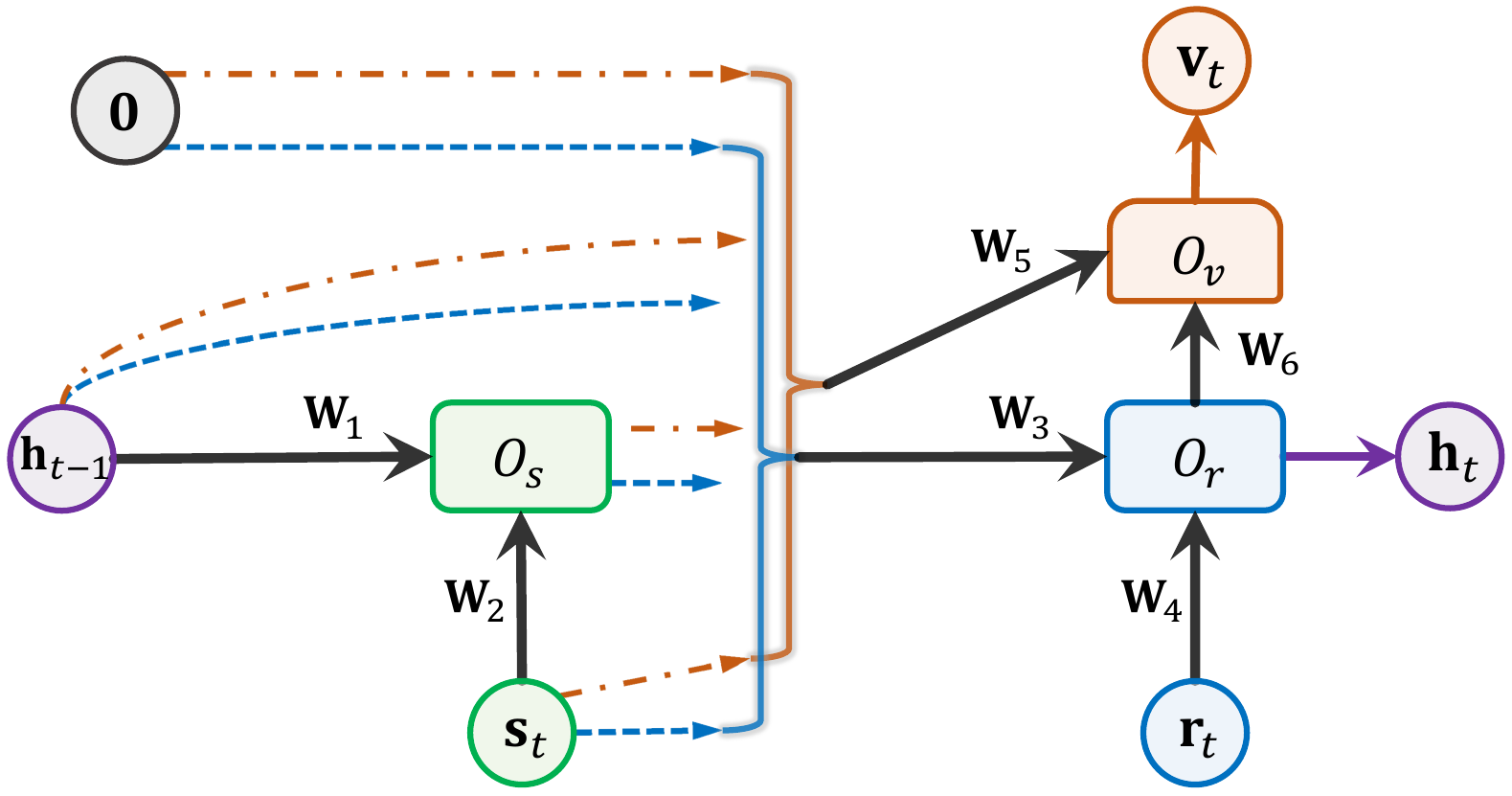}
	\caption{Search space $\mathcal{A}$ of $f$ for \eqref{eq:recurrent}.}
	\label{fig:struct}
	\end{minipage}
	\begin{minipage}{0.48\textwidth}
	\centering
	\captionsetup{type=table}
		\caption{The split search space of $f$ in Figure~\ref{fig:struct} into
			macro-level $\bm{\alpha}_1$ and micro-level $\bm{\alpha}_2$.}
	\label{tab:searchspace}
	\small
	\setlength\tabcolsep{2pt}
	\renewcommand{\arraystretch}{1.5}
	\vspace{-7px}
		\renewcommand{\arraystretch}{1.5}
	\begin{tabular}{c|C{50px}|C{90px}}
		\hline
				    macro-level     &   connections    & $\mathbf h_{t-1}, O_s, \mathbf  0, \mathbf s_t$ \\ \cline{2-3}
		$\bm{\alpha}_1 \in {\mathcal{A}_1}$ &   combinators    &  
		$+$, $\odot$, $\otimes$, gated 
		\\ \hline 
		    micro-level     &    activation    &             identity, tanh, sigmoid             \\ \cline{2-3}
$\bm{\alpha}_2 \in {\mathcal{A}_2}$ & weight matrix &               $\left\{ \mathbf{W}_i \right\}_{i = 1}^6$, $\mathbf I$                \\ \hline
	\end{tabular}
\end{minipage}
\vspace{-8px}
\end{figure}

In each step $t$,
we focus on one triplet $(s_t, r_t, o_t)$.
To design the search space $\mathcal A$,
we need to figure out what are important properties in \eqref{eq:recurrent}.
Since $\mathbf s_t$ and $\mathbf r_t$ 
are indexed from different embedding sets,
we introduce two operators to make a difference,
i.e. $O_s$  for $\mathbf s_t$ 
and $O_r$ for $\mathbf r_t$ 
as in Figure~\ref{fig:struct}.
Another operator $O_v$ is used to model the output $\mathbf v_t$ in \eqref{eq:recurrent}.
Then, the hidden state $\mathbf h_t$ is defined to propagate the information across triplets.
Taking Chains \cite{das2016chains} as an example,
$O_s$ is an adding combinator for $\mathbf h_{t-1}$ and $\mathbf s_t$,
$O_r$ is an RNN cell to combine $O_s$ and $\mathbf r_t$,
and $O_v$ directly outputs $O_r$  as $\mathbf v_t$.

To control the information flow,
we search the connections from input vectors to the outputs,
i.e. the dashed lines in Figure~\ref{fig:struct}.
Then,
the combinators, which combine two vectors into one, are important 
since they determine
how embedding are transformed,
e.g. ``+'' in TransE \cite{bordes2013translating} and
``$\odot$'' in DistMult \cite{yang2014embedding}.
As in the search space of RNN \cite{zoph2017neural}, 
we introduce activation functions 
\textit{tanh} and \textit{sigmoid}
to give non-linear squashing.
Each link is either a trainable weight matrix $\mathbf W$
or an identity matrix $\mathbf I$
to adjust the vectors.
Detailed information is listed in Table~\ref{tab:searchspace}.

Let the training and validation set be $\mathcal G_{\text{tra}}$ and $\mathcal G_{\text{val}}$,
$\mathcal{M}$ be the measurement on  $\mathcal G_{\text{val}}$ and
$\mathcal{L}$ be the loss on $\mathcal G_{\text{tra}}$.
To meet different requirements on $f$,
we propose to search the architecture $\bm \alpha$ of $f$ as a special RNN.
The network here is not a general RNN but one specific to the KG embedding tasks.
The problem is defined to find an architecture $\bm{\alpha}$
such that validation performance is maximized,
i.e.,
\begin{align}
\bm{\alpha}^*
= \arg\max\nolimits_{\bm \alpha \in \mathcal{A}} 
\mathcal{M} \left(f(\bm{F^*}; \bm \alpha), \mathcal{G}_{\text{val}}\right), 
 ~~~ \text{s.t.}  ~~~
\bm{F}^*= \arg\min\nolimits_{\bm F} \mathcal{L} \left( f(\bm{F};\bm \alpha), \mathcal{G}_{\text{tra}} \right),
\label{eq:searchprob}
\end{align}
which is a bi-level optimization problem and is non-trivial to solve.
First, the computation cost to get 
the optimal parameters
$\bm F^*$ is generally high.
Second, 
searching for $\bm\alpha \!\in\! \mathcal A$ 
is a discrete optimization problem \cite{liu2018darts}
and the space is 
large (in Appendix~\ref{app:space}).
Thus, 
how to efficiently search the architectures
is a big challenge.

Compared with standard RNNs, which recurrently model each input vectors,
the Interstellar models the relational path with triplets as basic unit.
In this way, we can determine 
how to model short-term information inside each triplet 
and what long-term information should be passed along the triplets.
This makes our search space distinctive for the KG embedding problems.

\subsection{Proposed Search Algorithm}
\label{sec:algorithm}

In Section~\ref{ssec:Interstellar}, 
we have introduced the search space $\mathcal A$,
which contains considerable different architectures.
Therefore,
how to
search efficiently in $\mathcal A$ is an important problem.
Designing appropriate optimization algorithm for the discrete architecture parameters
is a big challenge.

\subsubsection{Problems with Existing Algorithms}

As introduced in Section~\ref{ssec:rel:nas},
we can either choose the stand-alone approach 
or one-shot approach
to search the network architecture.
In order to search efficiently,
search algorithms should be designed for specific scenarios \cite{zoph2017neural,liu2018darts,akimoto2019adaptive}.
When the search space is complex, 
e.g. the macro space of CNNs 
and the tree-structured RNN cells \cite{zoph2017neural,pham2018efficient},
stand-alone approach is preferred 
since it can provide accurate evaluation feedback while PS scheme is not reliable \cite{bender2018understanding,sciuto2019evaluating}.
However, the computation cost of evaluating an architecture under the stand-alone approach is high,
preventing us to efficiently search in large spaces.

One-shot search algorithms are widely used in searching micro spaces,
e.g. cell structures in CNNs and simplified DAGs in RNN cells \cite{akimoto2019adaptive,pham2018efficient,liu2018darts}.
Searching architectures on a supernet with PS in the simplified space 
is relatively possible.
However,
since the search space in our problem is more complex 
and the embedding status influences the evaluation,
PS is not reliable (see Appendix~\ref{app:evalprob}).
Therefore, one-shot algorithms is not appropriate in our problem.

\subsubsection{Hybrid-search Algorithm}
\label{ssec:hybrid}
Even though the two types of algorithms  
have their limitations,
is it possible to take advantage from both of them?
Back to the development from stand-alone search in NASNet \cite{zoph2017neural}
to one-shot search in ENAS \cite{pham2018efficient},
the search space is simplified from a macro space to micro cells.
We are motivated to split the space $\mathcal A$ into 
a macro part $\mathcal{A}_1$
and a micro part $\mathcal{A}_2$ in Table~\ref{tab:searchspace}.
$\bm{\alpha}_1\in \mathcal{A}_1$ controls the connections and combinators, 
influencing information flow a lot;
and $\bm{\alpha}_2 \in \mathcal{A}_2$ fine-tunes the architecture through activations and weight matrix. 
Besides, we empirically observe that PS for $\bm{\alpha}_2$ is reliable (in Appendix~\ref{app:evalprob}).
Then we propose a hybrid-search method that can be both fast and accurate in Algorithm~\ref{alg:twostage}.
Specifically,
$\bm{\alpha}_1$ and $\bm{\alpha}_2$ are sampled from a controller $c$ ---
a distribution \cite{akimoto2019adaptive,xie2018snas}
or a neural network \cite{zoph2017neural}.
The evaluation feedback of $\bm{\alpha}_1$ and $\bm{\alpha}_2$ 
are obtained through the stand-along manner and one-shot manner respectively.
After the searching procedure, the best architecture is sampled and we fine-tune the hyper-parameters to 
achieve the better performance.

\begin{center}
	\vspace{-6px}
	\begin{algorithm}[ht]
		\caption{Proposed search recurrent architecture as the Interstellar algorithm.}
		\label{alg:twostage}
		\small
		\begin{algorithmic}[1]
			\REQUIRE search space $\mathcal{A} \equiv \mathcal{A}_1 \cup \mathcal{A}_2$ in Figure~\ref{fig:struct},
			controller $c$ for sampling 
			$\bm{\alpha} = \left[ \bm{\alpha}_2, \bm{\alpha}_1 \right]$.
			\REPEAT
			\STATE sample the \textit{micro-level} architecture
			$\bm{\alpha}_2\in\mathcal{A}_2$ by $c$;
			\STATE update the controller $c$ for $k_1$ steps using Algorithm~\ref{alg:standalone}
			(in \textit{stand-alone} manner);
			\STATE sample the \textit{macro-level} architecture $\bm{\alpha}_1\in\mathcal{A}_1$ by $c$;
			\STATE update the controller $c$ for $k_2$ steps using Algorithm~\ref{alg:oneshot}
			(in \textit{one-shot} manner);
			\UNTIL{termination}
			\STATE Fine-tune the hyper-parameters for the best architecture $\bm \alpha^* = [\bm{\alpha}_1^*, {\bm \alpha}_2^*]$ sampled from c.
			\label{step:finetune}
			\RETURN $\bm \alpha^*$ and the fine-tuned hyper-parameters.
		\end{algorithmic}
	\end{algorithm}
	\vspace{-8px}
\end{center}

In the macro-level (Algorithm~\ref{alg:standalone}),
once we get the architecture $\bm{\alpha}$,
the parameters are obtained by full model training.
This ensures that the evaluation feedback for macro architectures $\bm{\alpha}_1$ is reliable.
In the one-shot stage (Algorithm~\ref{alg:oneshot}),
the main difference is that,
the parameters $\bm F$ are not initialized
and different architectures are evaluated on the same set of $\bm F$,
i.e. by PS.
This improves the efficiency of evaluating micro architecture  $\bm{\alpha}_2$
without full model training.

\begin{center}
\vspace{-15px}
\begin{minipage}{0.50\textwidth}
	\begin{algorithm}[H]
		\caption{Macro-level ($\bm{\alpha}_1$) update}
		\label{alg:standalone}
		\small
		\begin{algorithmic}[1]
			\REQUIRE controller $c$, $\bm{\alpha}_2\in\mathcal{A}_2$, parameters $\bm F$.
			\STATE sample an individual $\bm{\alpha}_1$ by $c$ to get the
			architecture $\bm{\alpha}=[\bm{\alpha}_1, \bm{\alpha}_2]$;
			\STATE initialize $\bm F$ and train to obtain $\bm F^*$ until converge by minimizing  $L\left(f(\bm F, \bm \alpha), \mathcal G_{\text{tra}}\right)$;
			\STATE evaluate $\mathcal M(f(\bm F^*,  \bm{\alpha}), \mathcal G_{\text{val}})$ to update $c$.
			\label{step:sa:update} 
			\RETURN the updated controller $c$.
		\end{algorithmic}
	\end{algorithm}
\end{minipage}
\hfill
\begin{minipage}{0.47\textwidth}
	\begin{algorithm}[H]
		\caption{Micro-level ($\bm{\alpha}_2$) update}
		\label{alg:oneshot}
		\small
		\begin{algorithmic}[1]
			\REQUIRE controller $c$, $\bm{\alpha}_1\in\mathcal{A}_1$, parameters $\bm F$.
			\STATE sample an individual $\bm{\alpha}_2$ by $c$ to get the architecture $\bm{\alpha}=[\bm{\alpha}_1, \bm{\alpha}_2]$;
			\STATE  sample a mini-batch $\mathcal B_{\text{tra}}$ from $\mathcal G_{\text{tra}}$ and update $\bm F$ with gradient $\nabla_{\bm F}\mathcal L\left(f(\bm F, \bm \alpha), \mathcal B_{\text{tra}}\right)$; \label{step:os:f}
			\STATE sample a mini-batch $\mathcal B_{\text{val}}$ from $\mathcal G_{\text{val}}$ and evaluate $\mathcal M\left(f(\bm F, \bm{\alpha}), \mathcal B_{\text{val}}\right)$ to update $c$. \label{step:os:update}
			\RETURN the updated controller $c$.
		\end{algorithmic}
	\end{algorithm}
\end{minipage}
\vspace{-4px}
\end{center}

The remaining problem is how to model and update the controller,
i.e. step~\ref{step:sa:update} in Algorithm~\ref{alg:standalone}
and step~\ref{step:os:update} in Algorithm~\ref{alg:oneshot}.
The evaluation metrics in KG tasks are usually ranking-based, e.g. Hit@$k$,
and are non-differentiable.
Instead of using the direct gradient,
we turn to the derivative-free optimization methods \cite{conn2009introduction},
such as reinforcement learning 
(policy gradient in \cite{zoph2017neural} and Q-learning in \cite{baker2017designing}),
or Bayes optimization \cite{bergstra2011algorithms}.
Inspired by the success of policy gradient in searching CNNs and RNNs \cite{zoph2017neural,pham2018efficient},
we use policy gradient to optimize the controller $c$.

Following \cite{akimoto2019adaptive,geman1984stochastic,xie2018snas},
we use stochastic relaxation  for the architectures $\bm{\alpha}$.
Specifically, 
the parametric probability distribution $\bm \alpha\sim p_{\bm\theta}(\bm \alpha)$
is introduced on the search space $\bm \alpha\in\mathcal A$
and then the distribution parameter $\bm\theta$ is optimized to
maximize the expectation of validation performance
\begin{align}
\max_{\bm\theta}
J(\bm\theta) 
= 
\max_{\bm\theta}
\mathbb{E}_{\bm \alpha \sim p_{\bm\theta}(\bm \alpha)}
\left[
\mathcal{M} \left( f(\bm{F}^*;\bm{\alpha}), 
\mathcal{G}_{\text{val}}\right)
\right].
\label{eq:maxJ}
\vspace{-8px}
\end{align}
Then, the optimization target  is transformed from \eqref{eq:searchprob} with the discrete $\bm \alpha$  
into \eqref{eq:maxJ} with the continuous $\bm\theta$.
In this way, $\bm\theta$ is updated by $\bm\theta_{t + 1} \!=\! \bm\theta_t \!+\! \rho 
\nabla_{\bm\theta} J(\bm\theta_t)$,
where $\rho$ is the step-size,
 and 
$\nabla_{\bm\theta} J(\bm\theta) = \mathbb E\left[\mathcal M(f(F;\bm \alpha), \mathcal G_{\text{val}})\nabla_{\bm\theta}\ln\left(p_{\bm\theta_t}(\bm \alpha)\right)\right]$ is the pseudo gradient of $\bm \alpha$.
The key observation 
is that we do not need to take direct gradient w.r.t. $\mathcal{M}$,
which is not available.
Instead, 
we only need to get the validation performance measured by $\mathcal{M}$.

To further improve the efficiency,
we use natural policy gradient (NPG)~\cite{pascanu2013revisiting}
$\tilde{\nabla}_{\bm\theta} J(\bm\theta_t)$ to replace $\nabla_{\bm\theta} J(\bm\theta_t)$,
where $\tilde{\nabla}_{\bm\theta} J(\bm\theta_t)=\left[ \bm{H}\!\left(\bm\theta_t \right) \right]^{-1}
\nabla_{\bm\theta} J(\bm\theta_t)$ is computed by multiplying a Fisher information matrix $\bm{H} (\bm\theta_t )$ \cite{amari1998natural}.
NPG has shown to have better convergence speed \cite{akimoto2019adaptive,amari1998natural,pascanu2013revisiting} (see Appendix~\ref{ssec:ngimplement}).

\section{Experiments}
\vspace{-1px}
\subsection{Experiment Setup}
\vspace{-1px}

Following \cite{grover2016node2vec,guo2019learning,guu2015traversing},
we sample the relational paths from biased random walks (details in Appendix~\ref{app:path}).
We use two basic tasks in KG,
i.e. entity alignment and link prediction.
Same as the literature \cite{bordes2013translating,guo2019learning,wang2017knowledge,zhu2017iterative},
we use the ``filtered'' ranking metrics:
mean reciprocal ranking (MRR) and Hit@$k (k=1, 10)$.
Experiments are written in Python with PyTorch framework \cite{paszke2017automatic}
and run on a single 2080Ti GPU.
Statistics of the data set we use in this paper is in Appendix~\ref{app:datas}.
Training details of each task are given in Appendix~\ref{app:hyper}.
Besides, all of the searched models are shown in Appendix~\ref{app:models} due to space limitations.

\subsection{Understanding the Search Space}
\label{sec:case}
\vspace{-1px}
Here,
we illustrate the designed search space $\mathcal{A}$ in Section~\ref{ssec:Interstellar} 
using  Countries \cite{bouchard2015approximate} dataset,
which contains 271 countries and regions,
and 2 relations \textit{neighbor} and \textit{locatedin}.
This dataset contains three tasks:
S1 infers $neighbor \wedge locatedin \rightarrow locatedin$
or $locatedin \wedge locatedin \rightarrow locatedin$;
S2 require to infer the 2 hop relations $neighbor \wedge locatedin \rightarrow locatedin$;
S3 is harder and requires modeling 3 hop relations
$neighbor \wedge locatedin \wedge locatedin \rightarrow locatedin$.

To understand the dependency on the length of paths for various tasks,
we extract four subspaces from $\mathcal{A}$ (in Figure~\ref{fig:struct}) with different connections.
Specifically,
(P1) represents the single hop embedding models;
(P2) processes 2 successive steps;
(P3) models long relational paths without intermediate entities; and
(P4) includes both entities and relations along path.
Then we search $f$ in P1-P4 respectively to see how well these subspaces can tackle the tasks S1-S3.

\begin{figure}[ht]
	\centering
	\vspace{-8px}
	\subfigure[P1.]{\includegraphics[height = 2.6cm]{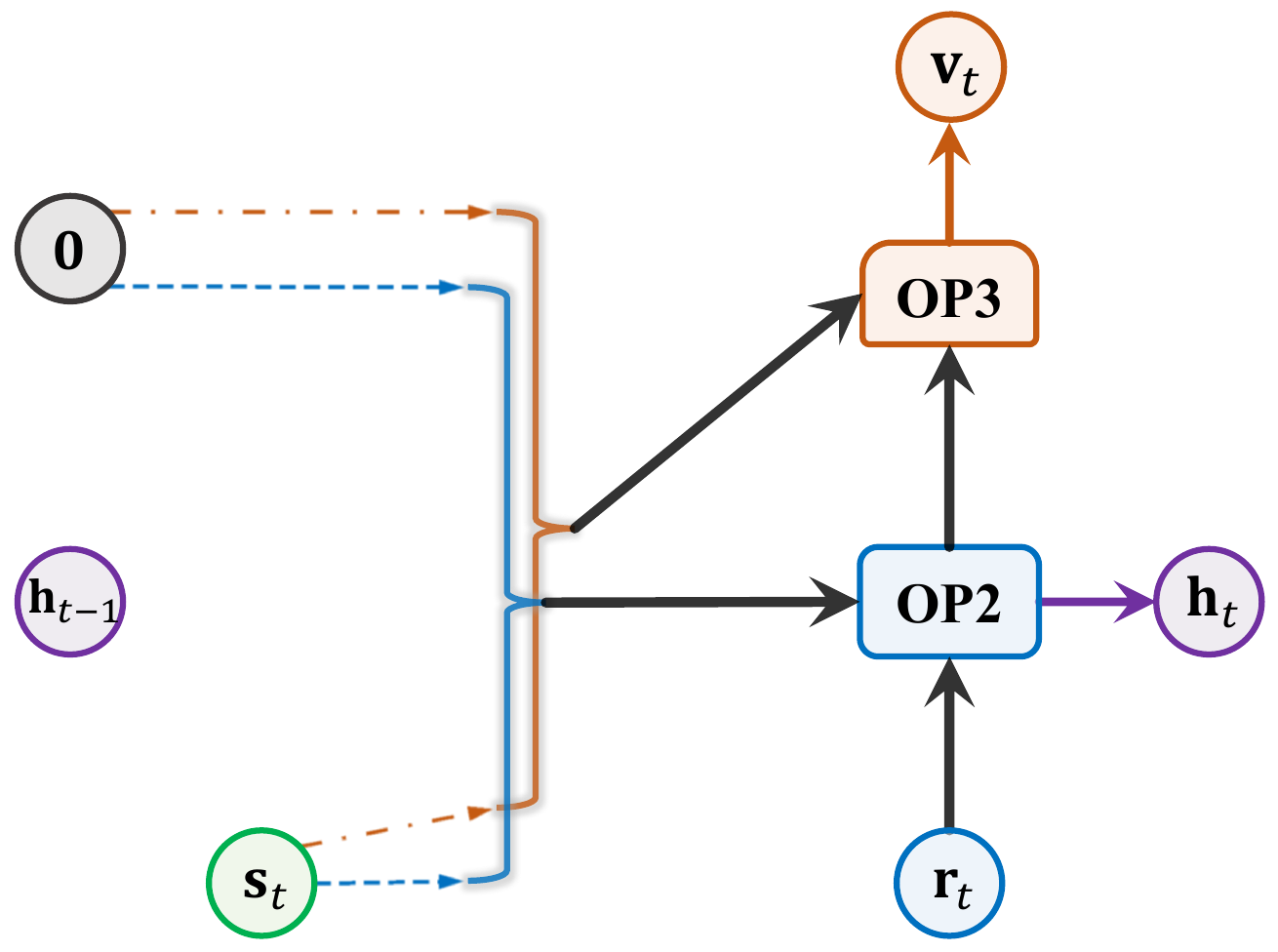}}
	\ 
	\subfigure[P2.]{\includegraphics[height = 2.6cm]{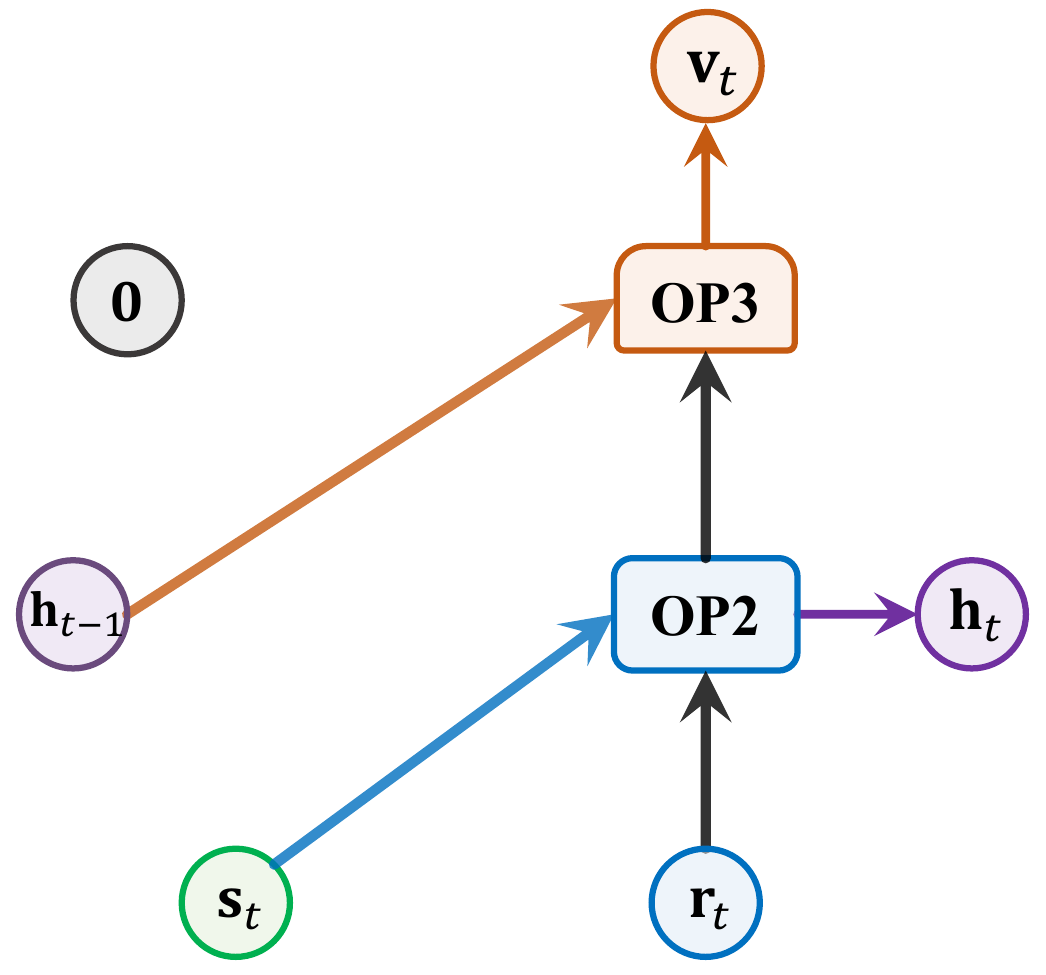}}
	\ 
	\subfigure[P3.]{\includegraphics[height = 2.6cm]{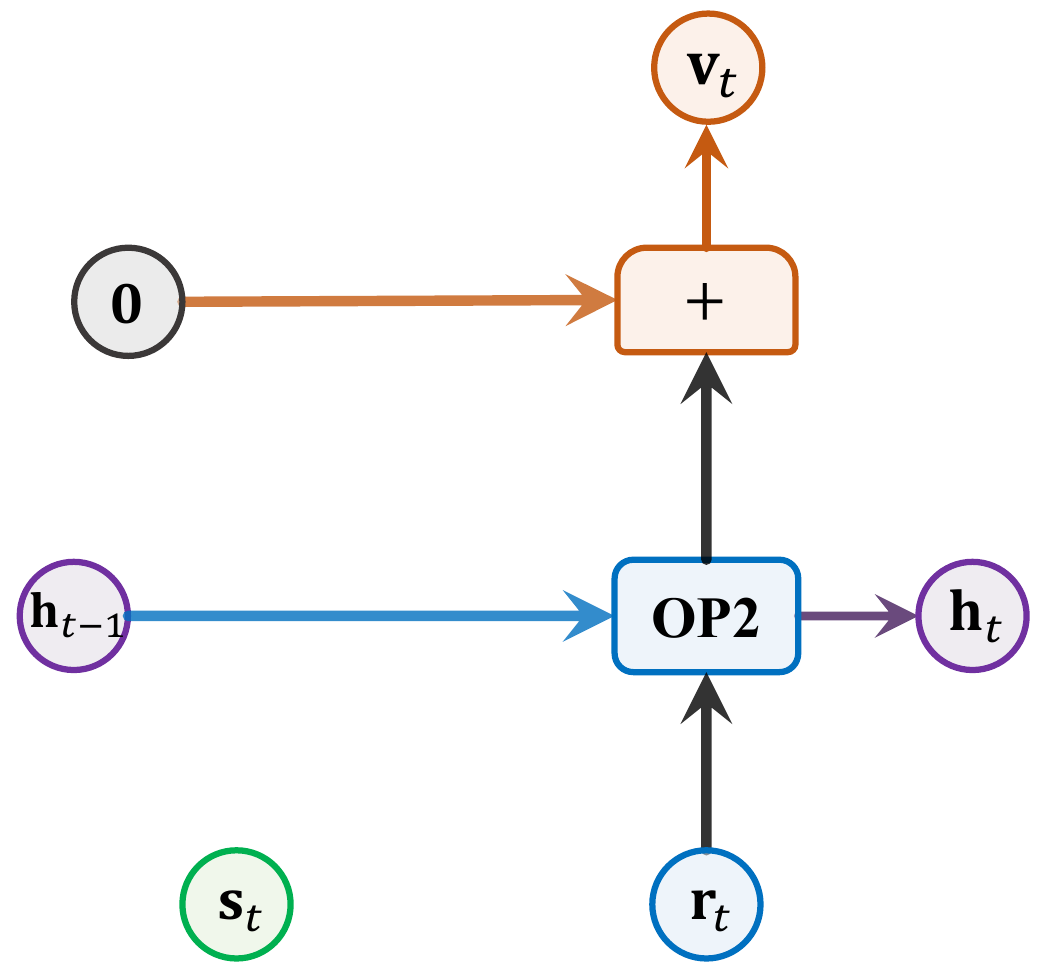}}
	\ 
	\subfigure[P4.]{\includegraphics[height = 2.6cm]{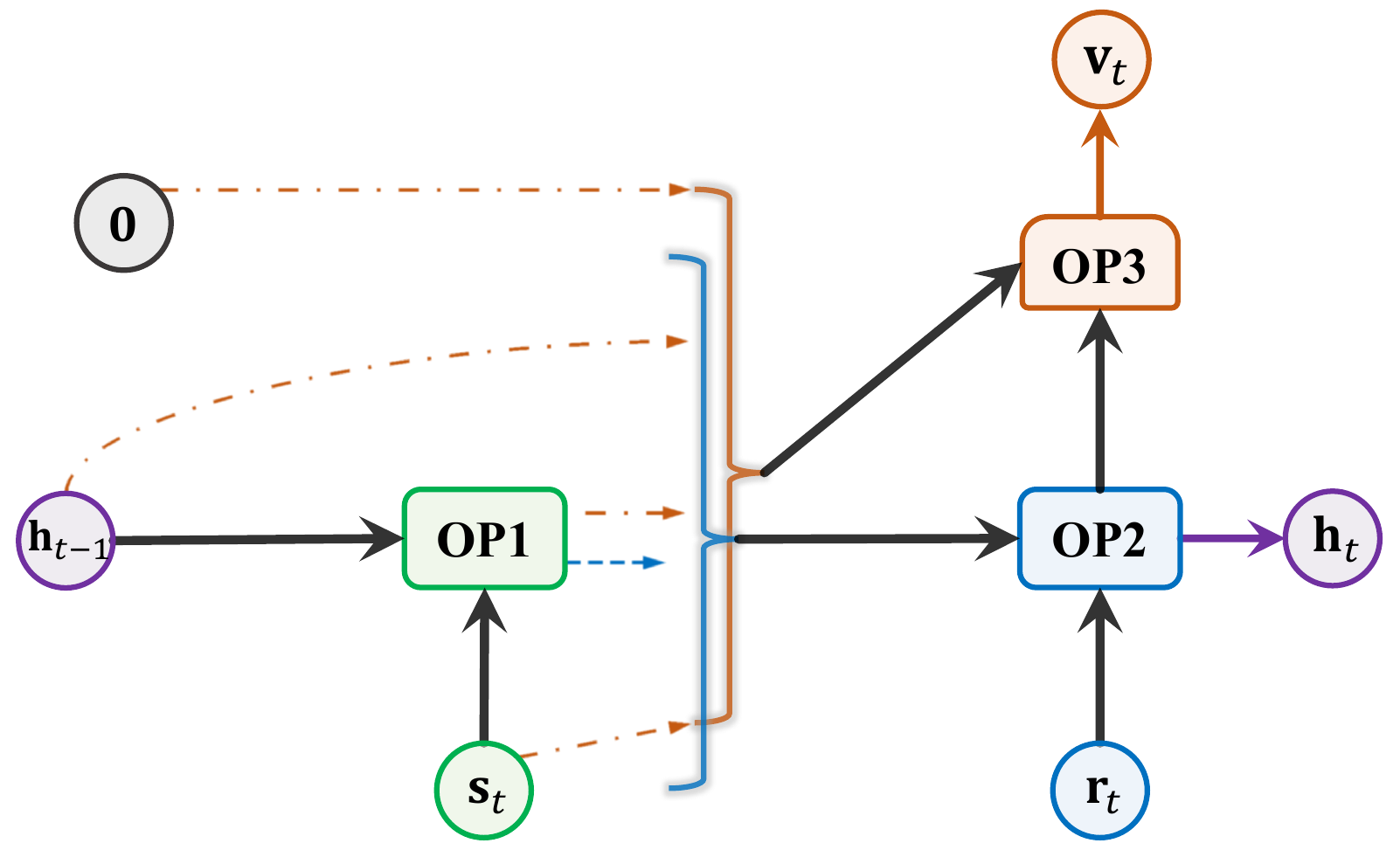}}
	\vspace{-10px}
	\caption{Four subspaces with different connection components in the recurrent search space.}
	\label{fig:countries}
	\vspace{-6px}
\end{figure}

For each task,
we randomly generate 100 models for each subspace
and record the model with
the best 
\textit{area under curve of precision recall} (AUC-PR) 
on validation set.
This procedure is repeated 5 times to evaluate the testing performance in Table~\ref{tab:countries}.
We can see that there is no single subspace
\begin{center}
	\vspace{-8px}
	\begin{minipage}{0.47\textwidth}
		 performing well on all tasks.
		For easy tasks S1 and S2,
		short-term information is more important.
		Incorporating entities along path like P4 is bad for learning 2 hop relationships.
		For the harder task S3,
		P3 and P4 outperform the others since it can model long-term information
		in more steps.
	\end{minipage}
\hfill
	\begin{minipage}{0.51\textwidth}
		\centering
		\vspace{-1px}
		\captionof{table}{Performance on Countries dataset.}
		\vspace{-8px}
		\label{tab:countries}
		\small
		\setlength\tabcolsep{2pt}
		\begin{tabular}{c|c|c|c}
			\hline
			&            S1            &            S2            &            S3            \\ \hline
			P1  &     0.998$\pm$0.001      &     0.997$\pm$0.002      &     0.933$\pm$0.031      \\
			P2  & \textbf{1.000$\pm$0.000} &     0.999$\pm$0.001      &     0.952$\pm$0.023      \\ 
			P3  &     0.992$\pm$0.001      & \textbf{1.000$\pm$0.000} &     0.961$\pm$0.016      \\ 
			P4  &     0.977$\pm$0.028      &     0.984$\pm$0.010      & \textbf{0.964$\pm$0.015} \\ \hline
			\textbf{Interstellar} & \textbf{1.000$\pm$0.000} & \textbf{1.000$\pm$0.000} &  \bf{0.968$\pm$ 0.007}   \\ \hline
		\end{tabular}
	\end{minipage}
\end{center}


Besides,
we evaluate the best model searched in the whole space $\mathcal A$ for S1-S3.
As in the last line of Table~\ref{tab:countries},
the model searched by Interstellar achieves good performance on the hard task S3.
Interstellar prefers different candidates (see Appendix~\ref{app:countries}) for S1-S3 over the same search space.
This verifies our analysis that it is difficult to use a unified model that can 
adapt to the short-term and long term information for different KG tasks.

%


\subsection{Comparison with State-of-the-art KG Embedding Methods}
\label{sec:comp:kge}

\textbf{Entity Alignment.} 
The entity alignment task aims to align entities in different KGs referring the same instance.
In this task,
long-term information is important since we need to propagate the alignment across triplets \cite{guo2019learning,sun2018bootstrapping,zhu2017iterative}.
We use four cross-lingual and cross-database subset from DBpedia
and Wikidata generated by \cite{guo2019learning},
i.e.
DBP-WD, DBP-YG, EN-FR, EN-DE.
For fair comparison, we follow the same path sampling scheme and the data set splits in \cite{guo2019learning}.


\begin{table}[ht]
	\centering
	\vspace{-12px}
	\caption{Performance comparison on entity alignment task. H@$k$ is short for Hit@$k$. The results of 
		TransD \cite{ji2015knowledge},
		BootEA \cite{sun2018bootstrapping}, 
		IPTransE \cite{zhu2017iterative},
		GCN-Align \cite{wang2018cross}
		and RSN \cite{guo2019learning}
		are copied from \cite{guo2019learning}.}
	\label{tab:ea:compare}
	\small
	\setlength\tabcolsep{2.75px}
	\begin{tabular}{cc|ccc|ccc|ccc|ccc}
\hline
		 \multicolumn{2}{c|}{\multirow{2}{*}{models}}   &          \multicolumn{3}{c|}{DBP-WD}          &          \multicolumn{3}{c|}{DBP-YG}          &          \multicolumn{3}{c|}{EN-FR}           &           \multicolumn{3}{c}{EN-DE}           \\ 
		                                &               &      H@1      &     H@10      &      MRR      &      H@1      &     H@10      &      MRR      &      H@1      &     H@10      &      MRR      &      H@1      &     H@10      &      MRR      \\ \hline
		   \multirow{3}{*}{triplet}    &    TransE     &     18.5      &     42.1      &     0.27      &      9.2      &     24.8      &     0.15      &     16.2      &     39.0      &     0.24      &     20.7      &     44.7      &     0.29      \\ 
		                                &    TransD*    &     27.7      &     57.2      &     0.37      &     17.3      &     41.6      &     0.26      &     21.1      &     47.9      &     0.30      &     24.4      &     50.0      &     0.33      \\ 
		                                &    BootEA*    &     32.3      &     63.1      &     0.42      &     31.3      &     62.5      &     0.42      &     31.3      &     62.9      &     0.42      &     44.2      &     70.1      &     0.53      \\ \hline
		    \multirow{3}{*}{GCN}      &   GCN-Align   &     17.7      &     37.8      &     0.25      &     19.3      &     41.5      &     0.27      &     15.5      &     34.5      &     0.22      &     25.3      &     46.4      &     0.22      \\ 
		    & VR-GCN &  19.4 	 &    55.5		& 0.32		& 20.9   & 55.7   & 0.32  & 16.0   & 50.8   &0.27   & 24.4  & 61.2  &   0.36 \\ 
		                                &     R-GCN     &     8.6    &    31.4     &  0.16     &  13.3   &   42.4  &  0.23   &     7.3    &   31.2   &  0.15   &   18.4        &    44.8   &  0.27    \\ \hline
		     \multirow{5}{*}{path}      &    PTransE    &     16.7      &     40.2      &     0.25      &      7.4      &     14.7      &     0.10      &      7.3      &     19.7      &     0.12      &     27.0      &     51.8      &     0.35      \\ 
		                                &   IPTransE*   &     23.1      &     51.7      &     0.33      &     22.7      &     50.0      &     0.32      &     25.5      &     55.7      &     0.36      &     31.3      &     59.2      &     0.41      \\ 
		                                &    Chains     &     32.2      &     60.0      &     0.42      &     35.3      &     64.0      &     0.45      &     31.4      &     60.1      &     0.41      &     41.3      &     68.9      &     0.51      \\ 
		                                &     RSN*      &     38.8      &     65.7      &     0.49      &     40.0      &     67.5      &     0.50      &     34.7      &     63.1      &     0.44      &     48.7      &     72.0      &     0.57      \\ \cline{2-14}
		                                & \textbf{Interstellar} & \textbf{40.7} & \textbf{71.2} & \textbf{0.51} & \textbf{40.2} & \textbf{72.0} & \textbf{0.51} & \textbf{35.5} & \textbf{67.9} & \textbf{0.46} & \textbf{50.1} & \textbf{75.6} & \textbf{0.59} \\ \hline
	\end{tabular}
\end{table}


Table~\ref{tab:ea:compare} compares the testing performance of the models searched by Interstellar and human-designed ones on the \textit{Normal} version datasets \cite{guo2019learning} 
(the \textit{Dense} version \cite{guo2019learning} in Appendix~\ref{app:EA:Dense}).
In general, the path-based models are better than 
the GCN-based and triplets-based models by modeling long-term dependencies.
BootEA \cite{sun2018bootstrapping} and IPTransE \cite{zhu2017iterative} win over TransE \cite{bordes2013translating} and PTransE \cite{lin2015modeling} respectively
by iteratively aligning discovered entity pairs.
Chains \cite{das2016chains} and RSN \cite{guo2019learning} outperform graph-based models and the other path-based models
by explicitly processing both entities and relations along path.
In comparison,
Interstellar is able to search and balance the short-term and long-term information adaptively,
thus gains the best performance.
We plot the learning curve on DBP-WD of some triplet, graph and path-based models in Figure~\ref{fig:EAcurve} to 
verify the effectiveness of the relational paths.


\begin{center}
\begin{minipage}{0.31\textwidth}
\textbf{Link Prediction.}
In this task, an incomplete KG
is given
and the target is to predict the missing entities in unknown links \cite{wang2017knowledge}.
We use three famous benchmark datasets, 
WN18-RR \cite{dettmers2017convolutional} and FB15k-237 \cite{toutanova2015observed},
which are more realistic than their superset WN18 and FB15k \cite{bordes2013translating},
and YAGO3-10 \cite{mahdisoltani2013yago3}, a much larger dataset.
\end{minipage}
\hfill
\begin{minipage}{0.67\textwidth}
\centering
\captionof{table}{Link prediction results.}
\setlength\tabcolsep{2pt}
\label{tab:lp:compare}
\small
\vspace{-8px}
\begin{tabular}{c|ccc|cccccc}
	\hline
	\multirow{2}{*}{models} &             \multicolumn{3}{c|}{WN18-RR}             &            \multicolumn{3}{c|}{FB15k-237}     &   \multicolumn{3}{c}{YAGO3-10}      \\ \cline{2-10}
	&H@1 &  H@10   & MRR     & H@1    &   H@10   & MRR    & H@1    &   H@10   & MRR \\ \hline
	TransE          &      12.5       &       44.5       &      0.18       &      17.3       &       37.9       &      0.24   &  10.3  &  27.9   &  0.16  \\ 
	ComplEx   &      41.4       &       49.0       &      0.44       &      22.7       &       49.5       &      {0.31}   & 40.5  &  62.8 & 0.48   \\ 
	RotatE* & 43.6 & 54.2 & {0.47}  & \textbf{23.3}  & 50.4  &  \textbf{0.32} &  40.2    &  63.1   & 0.48  \\ \hline
	R-GCN &   -     &     -    &   -    &     15.1    &   41.7    &      0.24     & -  & - & - \\ \hline
	PTransE    &      27.2       &       46.4       &      0.34       &      20.3       &       45.1       &      0.29    &  12.2  &  32.3   &    0.19 \\ 
	RSN           &      38.0       &       44.8       &      0.40       &      19.2       &       41.8       &      0.27    & 16.4  &  37.3   &    0.24   \\ \hline
	Interstellar           &     \textbf{43.8}      &    \textbf{54.6}    &  \textbf{0.48}     &     \textbf{23.3}    &    \textbf{50.8}    &    \textbf{0.32}  &  \textbf{42.4}   & \textbf{66.4}   & \textbf{0.51 }  \\ \hline
\end{tabular}
\end{minipage}
\vspace{-5px}
\end{center}

We search architectures with dimension $64$ to save time
and compare the models with dimension $256$.
Results of R-GCN on WN18-RR and YAGO3-10 are not available due to out-of-memory issue.
As shown in 
Table~\ref{tab:lp:compare},
PTransE outperforms TransE by modeling compositional relations, 
but worse than ComplEx and RotatE since the adding operation is inferior to $\otimes$
when modeling the interaction between entities and relations \cite{trouillon2017knowledge}.
RSN is worse than ComplEx/RotatE since it pays more attention to 
long-term information rather than the inside semantics.
Interstellar outperforms the path-based methods PTransE and RSN by searching architectures that model fewer steps 
(see Appendix~\ref{app:LPmodels}).
And it works comparable with the triplet-based models,
i.e. ComplEx and RotatE,
which are specially designed for this task.
We show the learning curve of TransE, ComplEx, RSN and Interstellar on WN18-RR in Figure~\ref{fig:LPcurve}.

\subsection{Comparison with Existing NAS Algorithms}
\label{sec:exp:auto}

\begin{figure}[ht]
	\vspace{-5px}
	\begin{minipage}{0.5\textwidth}
	\centering
	\subfigure[Entity alignment.]
	{\includegraphics[width=0.49\textwidth]{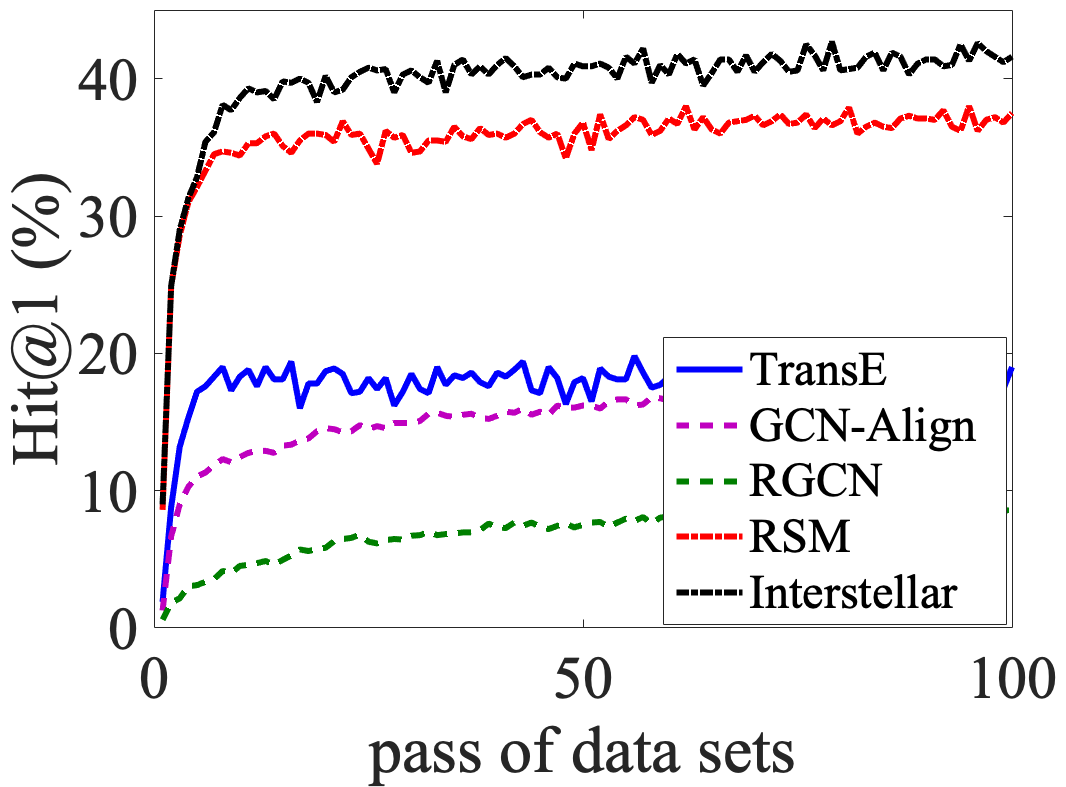}\label{fig:EAcurve}}
	\subfigure[Link prediction.]
	{\includegraphics[width=0.485\textwidth]{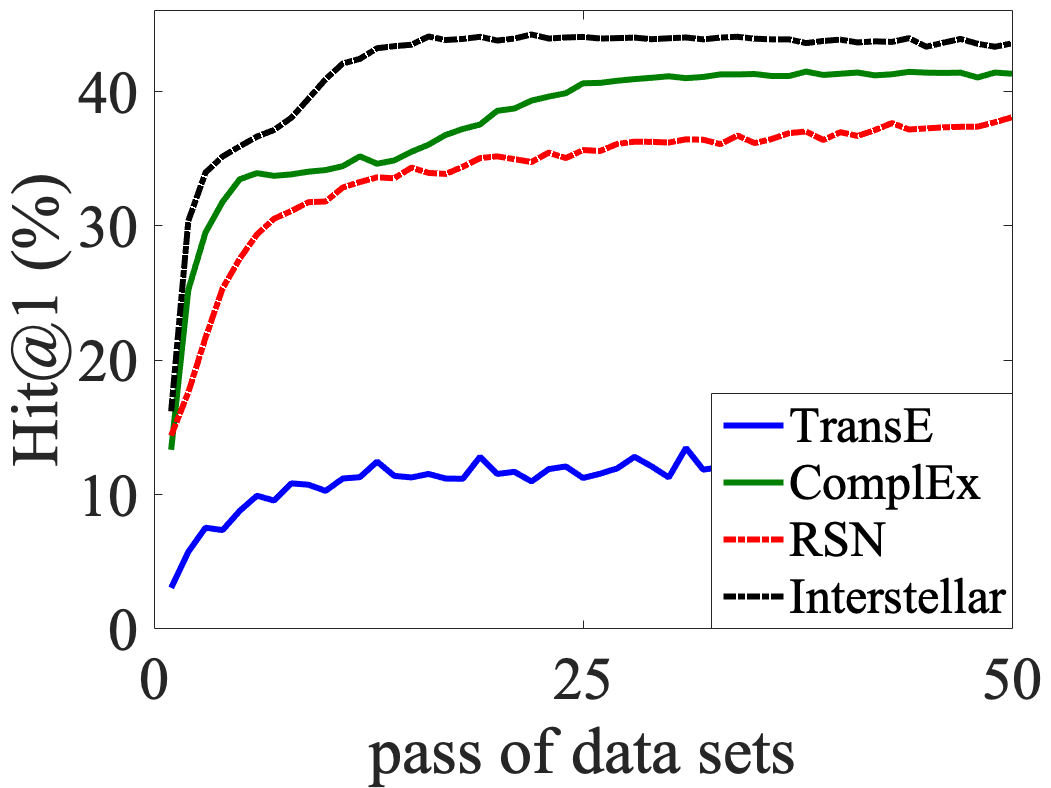}\label{fig:LPcurve}}
	\vspace{-10px}
	\caption{Single model learning curve. }
	\vspace{-5px}
	\label{fig:learingcurve}
	\end{minipage}
	\hfill
	\begin{minipage}{0.49\textwidth}
		\centering
		\subfigure[Entity alignment.]
		{\includegraphics[width=0.49\textwidth]{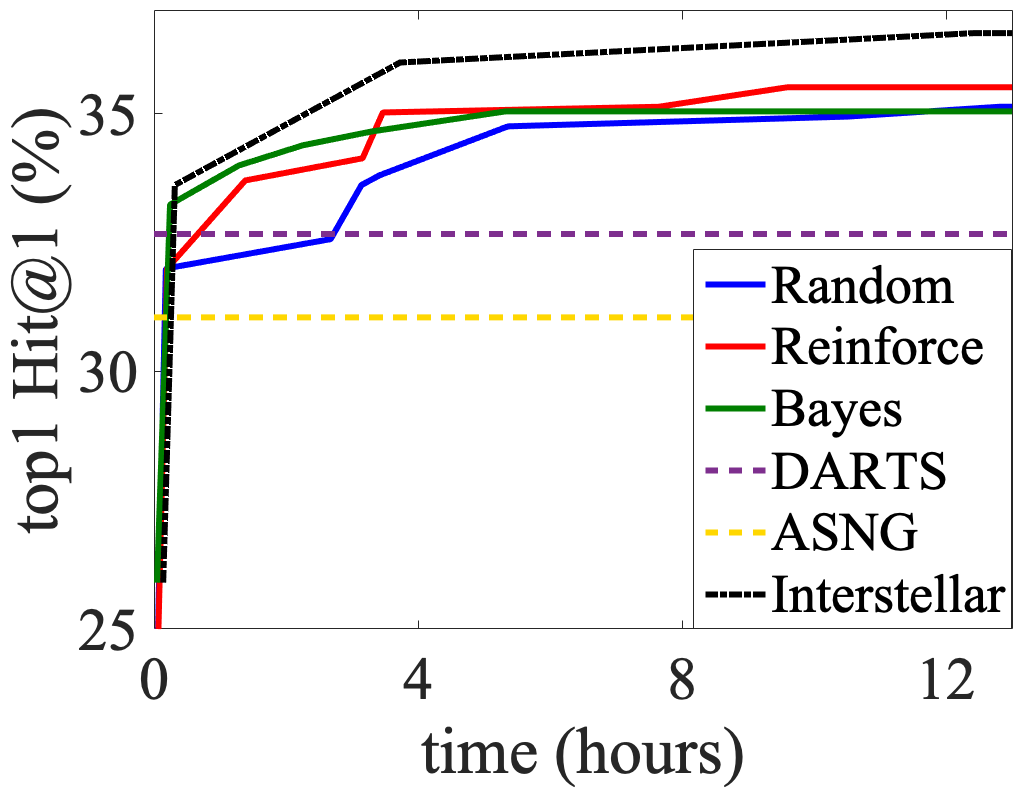}}
		\subfigure[Link prediction.]
		{\includegraphics[width=0.495\textwidth]{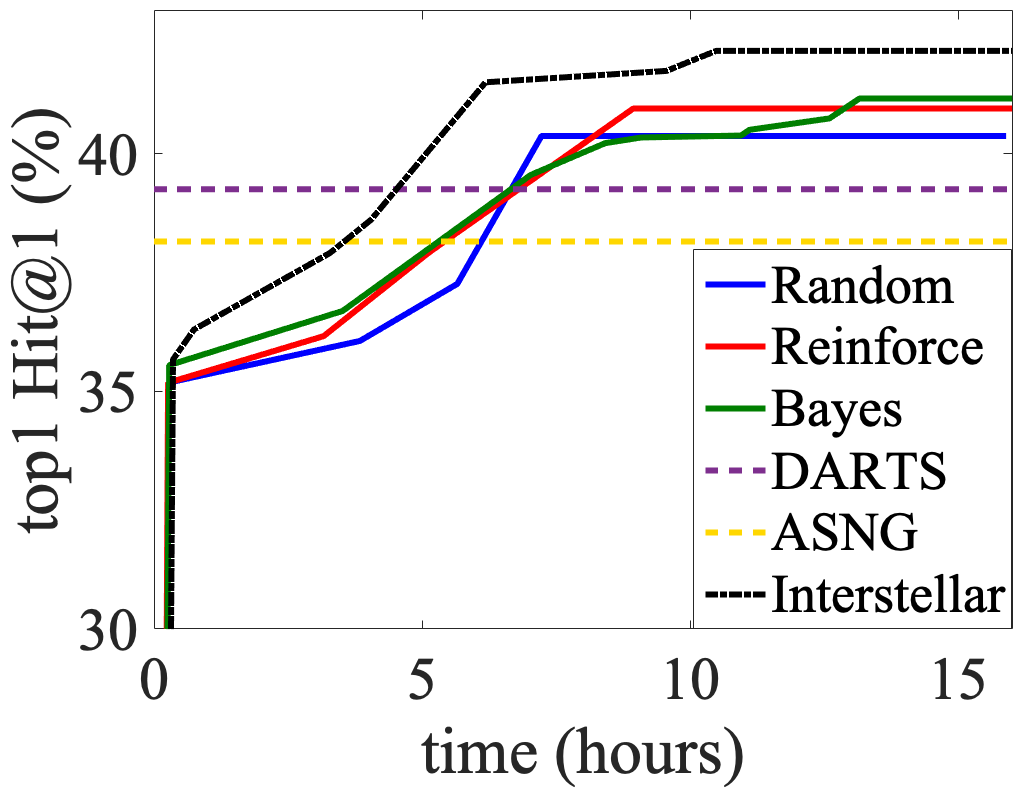}}
		\vspace{-10px}
		\caption{Compare with NAS methods.}
		\vspace{-5px}
		\label{fig:searchcurve}
	\end{minipage}
\vspace{-5px}
\end{figure}

In this part,
we compare the proposed Hybrid-search algorithm in Interstellar with the other NAS methods.
First, we compare the proposed algorithm with stand-alone NAS methods.
Once an architecture is sampled,
the parameters are initialized and trained into converge to give reliable feedback.
Random search \textit{(Random)}, Reinforcement learning \textit{(Reinforce)} \cite{zoph2017neural}
and Bayes optimization \textit{(Bayes)} \cite{bergstra2011algorithms}
are chosen as the baseline algorithms.
As shown in Figure~\ref{fig:searchcurve} (entity alignment on DBP-WD and link prediction on WN18-RR),
the Hybrid algorithm in Interstellar is more efficient 
since it takes advantage of the one-shot approach to search the micro architectures in $\mathcal{A}_2$.

Then,
we compare with one-shot NAS methods with PS on the entire space.
DARTS \cite{liu2018darts} and ASNG \cite{akimoto2019adaptive} are chosen as the baseline models.
For DARTS, the gradient is obtained from loss on training set
since validation metric is not differentiable.
We show the performance of the best architecture found by the two one-shot algorithms.
As shown in the dashed lines,
the architectures found by one-shot approach are much worse than that by the stand-alone approaches.
The reason is that PS is not reliable in our complex recurrent space (more experiments in Appendix~\ref{app:evalprob}).
In comparison, Interstellar is able to search reliably and efficiently
by taking advantage of both the stand-alone approach and the one-shot approach.




\label{sec:exp:abla}

\subsection{Searching Time Analysis}
\label{ssec:time:analy}
We show the clock time of Interstellar on entity alignment and link prediction tasks
in Table~\ref{tab:time:analy}.
Since the datasets used in entity alignment task have similar scales  
(see Appendix~\ref{app:datas}),
we show them in the same column.
For each task/dataset, we show the computation cost of
the macro-level and micro-level in Interstellar for 50 iterations between step 2-5 in Algorithm~\ref{alg:twostage} (20 for YAGO3-10 dataset);
and the fine-tuning procedure after searching for 50 groups of 
hyper-parameters,
i.e. learning rate, decay rate, dropout rate, L2 penalty and batch-size (details in Appendix~\ref{app:hyper}).
As shown, the entity alignment tasks take about 15-25 hours,
while link prediction tasks need about one or more days due to the larger data size.
The cost of the search process is at the same scale with that of the fine-tuning time,
which shows the search process is not expensive.

\begin{table}[ht]
	\centering
	\caption{Comparison of searching and fine-tuning time (in hours) in Algorithm~\ref{alg:twostage}.}
	\label{tab:time:analy}
	\small
	\setlength\tabcolsep{3pt}
	\begin{tabular}{cc|cc|ccc}
		\hline
		     \multicolumn{2}{c|}{\multirow{2}{*}{procedure}}       & \multicolumn{2}{c|}{entity alignment} &    \multicolumn{3}{c}{link prediction}    \\ \cline{3-7}
		       &                                                   &    Normal    &         Dense          &   WN18-RR    &         FB15k-237    &     YAGO3-10      \\ \hline
		\multirow{2}{*}{search} &              macro-level (line 2-3)               & 9.9$\pm$1.5  &      14.9$\pm$0.3      & 11.7$\pm$1.9 &        23.2$\pm$3.4     &   91.6$\pm$8.7  \\ 
		       &              micro-level (line 4-5)               & 4.2$\pm$0.2  &      7.5$\pm$0.6       & 6.3$\pm$ 0.9 &        5.6$\pm$0.4    &  10.4$\pm$1.3   \\ \hline
		\multicolumn{2}{c|}{fine-tune (line~\ref{step:finetune})} & 11.6$\pm$1.6 &      16.2$\pm$2.1      & 44.3$\pm$2.3 & 67.6$\pm$4.5  &  $>200$   \\ \hline
	\end{tabular}
	\vspace{-8px}
\end{table}

\section{Conclusion}

In this paper, we propose a new NAS method,
Interstellar,
to search RNN for learning from the relational paths,
which contain short-term and long-term information
in KGs.
By designing a specific search space based on
the important properties in relational path,
Interstellar can adaptively search promising architectures
for different KG tasks.
Furthermore,
we propose a hybrid-search algorithm that is more efficient
compared with the other the state-of-art NAS algorithms.
The experimental results verifies the effectiveness and efficiency of Interstellar on various KG embedding benchmarks.
In future work,
we can combine Interstellar with AutoSF \cite{zhang2020autosf} 
to give further improvement on the embedding learning problems.
Taking advantage of data similarity to improve the search efficiency on new datasets 
is another extension direction.

\clearpage
\section*{Broader impact}
Most of the attention on KG embedding learning has been focused on the triplet-based models.
In this work, we emphasis the benefits and importance of using relational paths to learn from KGs.
And we propose the path-interstellar as a recurrent neural architecture search problem.
This is the first work applying neural architecture search (NAS) methods on KG tasks. 

In order to search efficiently,
we propose a novel hybrid-search algorithm.
This algorithm addresses the limitations of stand-alone and one-shot search methods.
More importantly, the hybrid-search algorithm is not specific to the problem here. 
It is also possible to be applied to the other domains with more complex search space \cite{zoph2017neural, tan2019efficientnet,so2019evolved}.

One limitation of this work is that, the Interstellar is currently limited on the KG embedding tasks.
Extending to reasoning tasks like DRUM \cite{sadeghian2019drum} is an interesting direction.

\section*{Acknowledgment}
This work is partially supported by National Key Research and Development  Program of China Grant No. 2018AAA0101100,
the Hong Kong RGC GRF Project 16202218, CRF Project C6030-18G, C1031-18G, C5026-18G, 
AOE Project AoE/E-603/18, 
China NSFC No. 61729201, 
Guangdong Basic and Applied Basic Research Foundation 2019B151530001, 
Hong Kong ITC ITF grants ITS/044/18FX and ITS/470/18FX.
Lei Chen is partially supported by 
Microsoft Research Asia Collaborative Research Grant, 
Didi-HKUST joint research lab project,
 and Wechat and Webank Research Grants.

\bibliographystyle{plain}
\bibliography{bib}

\clearpage
\appendix

\section{Supplementary Details}

\subsection{Details of the search space}
\label{app:space}

\paragraph{Operators}
Given two input vectors $\mathbf a$ and $\mathbf b$ with $d$-dimensions and $d$ is an even number.
Two basic combinators are
(a). adding ($+$): $\mathbf o_i = \mathbf a_i+\mathbf b_i$;
and
(b).multiplying ($\odot$): $\mathbf o_i = \mathbf a_i \cdot \bm b_i$.
In order to cover ComplEx \cite{trouillon2017knowledge} in the search space,
we use Hermitian product ($\otimes$) as in \cite{trouillon2017knowledge}
\begin{equation*}
\mathbf o_i =\begin{cases}
\mathbf a_i \cdot \mathbf b_i - \mathbf a_{i+\nicefrac{d}{2}}\cdot \mathbf b_{i+\nicefrac{d}{2}} & \text{if } i<\nicefrac{d}{2} \\
\mathbf a_{i-\nicefrac{d}{2}} \cdot \mathbf b_{i} - \mathbf a_{i}\cdot \mathbf b_{i-\nicefrac{d}{2}} & \text{otherwise}
\end{cases}
\end{equation*}
As for the gated unit,
we define it as $\mathbf o_i = \mathbf g_i\cdot\mathbf a_i + (1-\mathbf g_i)\cdot \mathbf b_i$,
where the gate function $\mathbf g = \text{sigmoid}\left(\mathbf W_a \mathbf a + \mathbf W_b \mathbf b \right)$
with trainable parameters $\mathbf W_a, \mathbf W_b\in\mathbb R^{d\times d}$,
to imitate the gated recurrent function.

\paragraph{Space size}
Based on the details in Table~\ref{tab:searchspace},
the inputs of $O_r$ and $O_v$ are chosen from $\mathbf h_{t-1}, O_s, \mathbf 0, \mathbf s_t$.
Then the three operators $O_s, O_r, O_v$ select one combinator from $+, \odot, \otimes$ and gated.
Thus, we have $4^2 \times 4^3 = 1024$ architectures in the macro space $\mathcal{A}_1$.
For the micro space,
we only use activation functions after $O_s, O_r$ 
since we empirically observe that activation function on $O_v$ is bad for the embedding spaces.
In this way, the size of micro space $\mathcal{A}_2$ is $3^2\times 2^6=576$.
Totally, there should be $6\times 10^5$ candidates with the combination of macro-level space and micro-level space.


\subsection{Details of the search algorithm}
\label{ssec:ngimplement}

\paragraph{Compare with existing NAS problem.}
In Section~\ref{sec:algorithm}, we have already talked about the problems of existing search algorithms
and propose a hybrid-search algorithm that can search fast and accurate.
Here,
we give an overview of the different NAS methods in 
Table~\ref{tab:diff}.

\begin{table}[ht]
	\centering
	\caption{Comparison of state-of-the-art NAS methods for general RNN with the proposed Interstellar.}
	\small
	\begin{tabular}{c| C{105px}  C{105px}|c}
		\toprule
		&                                                                              \multicolumn{2}{c|}{Existing NAS Algorithms}                                                                               & \multirow{2}{*}{Interstellar} \\ 
		\multicolumn{1}{c|}{} & \multicolumn{1}{c}{Stand-alone approach}                                                    & \multicolumn{1}{c|}{One-shot approach}                                                                    &                        \\ \midrule
		space         &                        complex structures                         &                            micro cells                                            &      KG-specific       \\ \midrule
		algorithm       & reinforcement learning \cite{zoph2017neural}, Bayes optimization \cite{bergstra2011algorithms,kandasamy2018neural} & direct gradient descent \cite{liu2018darts}, stochastic relaxation \cite{akimoto2019adaptive,xie2018snas} &    natural policy gradient    \\ \midrule
		evaluation       & full model training                                                                         & parameter-sharing                                                                                         &         hybrid         \\ \bottomrule
	\end{tabular}
	\label{tab:diff}
\end{table}

\paragraph{Implementation of $\mathcal M(f(\bm F^*,  \bm{\alpha}), \mathcal G_{\text{val}})$.}
For both tasks, the training samples are composed of relational paths. But the validation data format is different.
For the entity alignment task, the validation data is a set of entity pairs,
i.e. $\mathcal S_{val} = \{(e_1, e_2)|e_1\in\mathcal E_1, e_2\in\mathcal E_2\}$.
The performance is measured by the cosine similarity of the embeddings $\mathbf e_1$ and $\mathbf e_2$.
Thus, the architecture $\bm \alpha$ is not available in $\mathcal M(f(\bm F^*,  \bm{\alpha}), \mathcal G_{\text{val}})$.
Instead, we use the negative loss function on a training mini-batch $-\mathcal L(f(\bm F^*,  \bm{\alpha}), \mathcal B_{\text{tra}})$ as an alternative of $\mathcal M(f(\bm F^*,  \bm{\alpha}), \mathcal G_{\text{val}})$.
For the link prediction task, the validation data is set of single triplets, i.e. 1-step paths.
Different from the entity alignment task, the path is available for validation measurement here.
Therefore, once we sample a mini-batch $\mathcal B_{\text{val}}$ from $\mathcal G_{\text{val}}$,
we can either use the loss function $-\mathcal L(f(\bm F^*,  \bm{\alpha}), \mathcal B_{\text{val}})$  
or the direct evaluation metric (MRR or Hit@$k$) $\mathcal M(f(\bm F^*,  \bm{\alpha}), \mathcal B_{\text{val}})$
on the mini-batch $\mathcal B_{\text{val}}$.


\paragraph{Implementation of natural policy gradient.}
Recall that the optimization problem of architectures is changed from optimizing $\bm \alpha$ by
$\bm{\alpha}^*
= \arg\max\nolimits_{\bm \alpha \in \mathcal{A}} 
\mathcal{M} \left(f(\bm{F^*}; \bm \alpha), \mathcal{G}_{\text{val}}\right)$
to optimizing the distribution parameter $\bm\theta$ by 
$\max_{\bm\theta} J(\bm\theta) = \max_{\bm\theta}
\mathbb{E}_{\bm \alpha \sim p_{\bm\theta}(\bm \alpha)}
\left[
\mathcal{M} \left( f(\bm{F}^*;\bm{\alpha}), 
\mathcal{G}_{\text{val}}\right)
\right].$
As discussed in Section~\ref{ssec:hybrid},
we use the natural policy gradient to solve $\bm\theta^* = \arg\max_{\bm\theta} J(\theta)$. 
To begin with, we need to define the distribution $p_{\bm\theta}$.
Same as \cite{akimoto2019adaptive,ollivier2017information}, we refer to the exponential family
$h(\bm{\alpha})\cdot \exp\left(\eta(\bm\theta)^\top T(\bm{\alpha})-A(\bm\theta)\right)$,
where $h(\bm{\alpha})$, $T(\bm{\alpha})$ and $A(\bm\theta)$ are known functions depending on the target distribution.
The benefit of using exponential family is that the inverse Fisher information matrix 
$\bm H^{-1}\left(\bm\theta\right)$ is easily obtained.
For simplicity, we set $h(\bm{\alpha})=1$
and choose the expectation parameters $\bm\theta = \mathbb E_{p_{\bm\theta}}[T(\bm{\alpha})]$ as in \cite{guo2019learning}.
Then the gradient reduces to $\nabla_{\bm\theta}\ln(p_{\bm\theta}(\bm\alpha)) = T(\bm{\alpha}) - \bm\theta$,
and the inverse Fisher information matrix
$\bm H^{-1}(\bm\theta) = \mathbb E_{p_{\bm\theta}}[\left(T(\bm{\alpha}) - \theta\right)\left(T(\bm{\alpha}) - \theta\right)^\top ]$
is computed with the finite approximation. The iteration is,
\[\bm\theta_{t +1} = \bm\theta_t + \rho \frac{1}{m}\sum_{i=1}^m \left[\left(T({\bm\alpha^{(i)}})-\bm\theta_t\right) \left(T({\bm\alpha^{(i)}})-\bm\theta_t\right)^\top\cdot\mathcal M\left(f({\bm F^{(i)}}^*; \bm \alpha^{(i)}), \mathcal G_{\text{val}}\right)\right],\]
where $\bm\alpha^{(i)}$'s are sampled from $p_{\bm\theta_t}$.
In this paper, we set the finite sample $m$ as 2 to save time.
Since the architecture parameters are categorical,
we model $p_{\bm\theta}$ as categorical distributions.
For a component with $K$ categories, 
we use a $K$ dimensional vector $\bm\theta$ to model the distribution $p_{\bm\theta}$.
The probability of sampling the $i$-th category is $\bm\theta_i$ and $\theta_K = 1-\sum_{j=1}^{K-1}\bm\theta_j$.
Note that, the NG is only used to update the first $K-1$ dimension of $\bm\theta$, and
$T(\bm\alpha_i)$ is a one-hot vector (without dimension $K$) of the category $\bm \alpha_i$.
For more details, please refer to Section~2.4 in \cite{guo2019learning} and the published code.

\subsection{Training details}
\label{app:path}

\begin{figure}[ht]
	\centering
	\includegraphics[height=2.6cm]{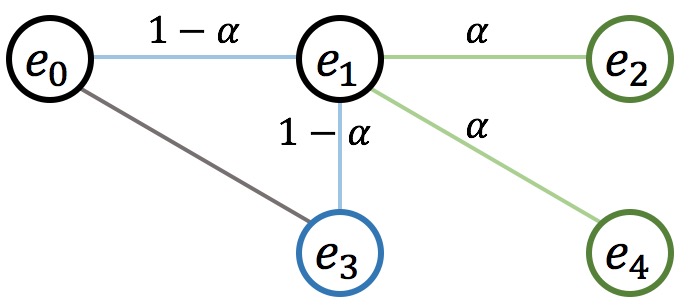}
	\caption{An example of the biased random walks.}
	\label{fig:walks}
\end{figure}

Following \cite{grover2016node2vec,guo2019learning},
we sample paths from biased random walks.
Take Figure~\ref{fig:walks} as an example 
and the path walked from $e_0$ to $e_1$ just now.
In conventional random walk,
all the neighbors of $e_1$
have the same probability to be sampled as the next step.
In biased random walk, we sample the neighbor that can go deeper or go to another KG with larger probability.
For single KG,
we use one parameter $\alpha\in(0.5, 1)$ to give the depth bias. 
Specifically,
the neighbors that are two steps away
from the previous entity $e_0$,
like $e_2$ and $e_4$ in Figure~\ref{fig:walks},
have the bias of $\alpha$.
And the bias of others, i.e. $e_0,$ $e_3$,
are $1-\alpha$.
Since $\alpha>0.5$, the path is more likely to go deeper. 
Similarly,
we have another parameter $\beta\in(0.5, 1)$ for two KGs to give cross-KG bias.
Assume $e_0,$ $e_1,$ and $e_3$ belongs to the same KG $\mathcal G_1$
and $e_2,$ $e_4$ are parts of $\mathcal G_2$.
Then, the next step is encouraged to jump to entities in another KG,
namely to the entities $e_2$ and $e_4$ in Figure~\ref{fig:walks}.
Since the aligned pairs are usually rare in the training set,
encouraging cross-KG walk can learn more about the aligned entities in the two KGs.

Since the KGs are usually large,
we sample two paths for each triplet in $\mathcal S_{tra}$.
The length of paths is 7 for entity alignment task on two KGs
and 3 for link prediction task on single KG.
The paths are fixed for each dataset to keep a fair comparison among different models.
The parameters used for sampling path are summarized in Table~\ref{tab:pathparam}.

\begin{table}[ht]
	\centering
	\caption{Parameter used for sampling path.}
	\label{tab:pathparam}
	\renewcommand{\arraystretch}{1.2}
	\begin{tabular}{c|c|c|c}
		\hline
		parameters    & $\alpha$ & $\beta$ & length \\ \hline
		entity alignment &   0.9    &   0.9   &   7    \\ \hline
		link prediction  &   0.7    &   --    &   3    \\ \hline
	\end{tabular}
\end{table}

Once the paths are sampled, we start training based on the relational path.
The loss for a path with length $L$, 
i.e. containing $L$ triplets,
is given in given as
\begin{equation} 
\mathcal L_{tra} = \sum_{t=1}^{L}\left\{-\mathbf v_t\cdot\mathbf o_t + \log\left(\sum_{o_i\in\mathcal E}\exp\left(\mathbf v_t\cdot\mathbf o_i\right)\right) \right\}
\label{eq:loss}
\end{equation}
In the recurrent step $t$, we focus on one triplet $(s_t, r_t, o_t)$.
The subject entity embedding $\mathbf s_t$ and relation embedding $\mathbf r_t$
are processed along with the proceeding information $\mathbf h_{t-1}$ to get the output $\mathbf v_t$.
The output $\mathbf v_t$ is encouraged to approach the object entity embedding $\mathbf o_t$.
Thus, the objective  in \eqref{eq:loss} can be regarded as a multi-class log-loss \cite{lacroix2018canonical}
where the object $o_t$ is the true label.

Besides, the training of the parameters $\bm F$ is based on mini-batch gradient descent.
Adam \cite{kingma2014adam} is used as the optimization method for updating model parameters.

\subsection{Data statistics}
\label{app:datas}

The datasets used in entity alignment task are the customized version by \cite{guo2019learning}.
For the normal version data,
nodes are sampled to approximate the degree distribution of original KGs.
This makes the datasets in this version more realistic.
For the dense version data,
the entities with low degrees are randomly removed in the original KGs.
This makes the data more similar to those used by existing methods \cite{sun2018bootstrapping,wang2018cross}.
We give the statistics in Table~\ref{tab:data:ea}.

\begin{table}[H]
	\centering
	\caption{Statistics of the datasets we used for entity alignment. 
		Each single KG contains 15,000 entities.
		There are 4,500 aligned entity pairs in the training sets,
		and 11,500 pairs for evaluation. 
		The first 10\% pairs among the 11,500 are used for validation and the left for testing. 
		We use ``\#'' short for ``number of''.}
	\label{tab:data:ea}
	\small
		\renewcommand{\arraystretch}{1.2}
	\begin{tabular}{c|c|cc|cc|cc|cc}
		\hline
		 \multirow{2}{*}{$\!\!$version$\!\!$} &  data   & \multicolumn{2}{c|}{DBP-WD} & \multicolumn{2}{c|}{DBP-YG} & \multicolumn{2}{c|}{EN-FR} & \multicolumn{2}{c}{EN-DE} \\ 
		 & source & $\!$DBpedia$\!$        & $\!$Wikidata$\!$       & $\!$DBpedia$\!$       & $\!$YAGO3$\!$       & $\!$English$\!$      & $\!$French$\!$       & $\!$English$\!$      & $\!$German$\!$      \\ \hline 
		\multirow{2}{*}{$\!\!$Normal$\!\!$} & $\!\!$\# relations$\!\!$ &       253         &    144        &    219           &     30        &     211         &   177          &      225        &     118        \\ 
		& $\!\!$\# triplets$\!\!$         &     38,421       &      40,159         &     33,571        &      34,660        &      36,508       &  33,532       &           38,281  &   37,069   \\ \hline
		\multirow{2}{*}{$\!\!$Dense$\!\!$} & $\!\!$\# relations$\!\!$ &       220         &    135        &    206           &     30        &     217         &   174          &      207        &     117        \\ 
		& $\!\!$\# triplets$\!\!$        &     68,598      &      75,465         &     71,257        &      97,131        &      71,929       &  66,760       &           56,983  &   59,848   \\ \hline
	\end{tabular}
	\label{tab:data:entity}
\end{table}

The datasets used for the link prediction task are used in many baseline models \cite{dettmers2017convolutional,sun2019rotate,guo2019learning}.
The WN18-RR and FB15k-237 datasets are created so that the link leakage problem \cite{dettmers2017convolutional,trouillon2017knowledge} can be solved.
Thus,
they are more realistic than their super-set WN18 and FB15k \cite{bordes2013translating}.

\begin{table}[H]
	\centering
	\caption{Statistics of the datasets used for link prediction.}
	\label{tab:data:lp}
	\renewcommand{\arraystretch}{1.1}
	\begin{tabular}{c|ccccc}
		\toprule
		Dataset & \#entity & \#relations & \#train & \#valid & \#test \\ 
		\midrule
		WN18-RR & 40,943 & 11 & 86,835 & 3,034 & 3,134 \\
		FB15k-237 & 14,541 & 237 & 272,115 & 17,535 & 20,466 \\
		YAGO3-10 & 123,188  & 37 & 1,079,040  & 5,000 & 5,000 \\ 
		\bottomrule 
	\end{tabular}
\end{table}

\subsection{Hyper-parameters and Training Details}
\label{app:hyper}

During searching, the meta hyper-parameters $k_1$ is one step 
and $k_2$ is one epoch based on the efficiency consideration.

In order to find the hyper-parameters during search procedure,
we use RSN \cite{guo2019learning} as a standard baseline to tune the hyper-parameters.
We use the \textit{HyperOpt} package with tree parsen estimator \cite{bergstra2011algorithms}
to search the 
learning rate $\eta$, L2 penalty $\lambda$, decay rate $u$,
batch size $m$,
as well as a dropout rate $p$.
The decay rate is applied to the learning rate each epoch
and the dropout rate is applied to the input embeddings.
The tuning ranges are given in Table~\ref{tab:hyper}.

\begin{table}[H]
	\centering
	\caption{Searching range of hyper-parameters}
	\renewcommand{\arraystretch}{1.3}
	\label{tab:hyper}
	\begin{tabular}{c|c}
		\hline
		hyper-param & ranges \\ \hline
		$\eta$         &    $[10^{-5}, 10^{-3}]$    \\ \hline
		$\lambda$      &   $[10^{-5}, 10^{-2}]$     \\ \hline
		$u$           &   $[0.98, 1]$     \\ \hline
		$m$           &    $\{128, 256, 512, 1024, 2048\}$    \\ \hline
		$p$           &   $[0, 0.6]$      \\ \hline
	\end{tabular}
\end{table}

The embedding dimension for entity alignment task is 256, 
and for link prediction is 64 during searching.
After the hyper-parameter tuning, 
the proposed Interstellar starts searching with this hyper-parameter setting.
For all the tasks and datasets, we search and evaluate 100 architectures.
Among them, we select the best architecture indicated by the Hit@1 performance on validation set,
i.e. the architecture with top~1 performance in the curve of Interstellar in Figure~\ref{fig:searchcurve}.
After searching,
the embedding size on link prediction is increased to be 256 and
we fine-tune the other hyper-parameters in Table~\ref{tab:hyper} again for the searched architectures.
Finally, the performance on testing data is reported in Table~\ref{tab:ea:compare}.
The time cost of searching and fine-tuning is given in Table~\ref{tab:time:analy} in Section~\ref{ssec:time:analy}.



\section{Supplementary Experiments}

\subsection{Entity alignment on the Dense version}
\label{app:EA:Dense}

Table~\ref{tab:ea:dense} compares the testing performance of the models searched by Interstellar
and human-designed ones on the \textit{Dense} version datasets \cite{guo2019learning}.
In this version, 
there is no much gap between the triplet-based and GCN-based models
since more data  can be used for learning the triplet-based ones.
In comparison, the path-based models are better by traversing across two KGs.
Among them,
Interstellar performs the best on DBP-WD, EN-FR and EN-DE and is comparable with RSN on DBP-YG dataset.
Besides, the searched architectures are different between the Normal version and Dense version,
as will be illustrated Appendix~\ref{app:EAmodels}.

\begin{table}[H]
	\centering
	\caption{Performance comparison on entity alignment tasks on dense dataset. H@$k$ is short for Hit@$k$. The results of 
		TransD \cite{ji2015knowledge},
		BootEA \cite{sun2018bootstrapping}, 
		IPTransE \cite{zhu2017iterative},
		GCN-Align \cite{wang2018cross}
		and RSN \cite{guo2019learning}
		are copied from \cite{guo2019learning}.}
	\label{tab:ea:dense}
	\small
	\setlength\tabcolsep{2.75px}
	\begin{tabular}{cc|ccc|ccc|ccc|ccc}
		\toprule
		\multicolumn{2}{c|}{\multirow{2}{*}{models}}   &          \multicolumn{3}{c|}{DBP-WD}          &          \multicolumn{3}{c|}{DBP-YG}          &          \multicolumn{3}{c|}{EN-FR}           &           \multicolumn{3}{c}{EN-DE}           \\ 
		&               &      H@1      &     H@10      &      MRR      &      H@1      &     H@10      &      MRR      &      H@1      &     H@10      &      MRR      &      H@1      &     H@10      &      MRR      \\ \midrule
		\multirow{3}{*}{triplet}    &    TransE     &     54.9    &    75.5    &  0.63  &     33.8      &   68.5     &  0.46      &   22.3   &  39.8    &  0.28  &    30.2       &       60.2        &      0.40     \\ 
		&    TransD*   &     60.5      &     86.3      &     0.69      &     62.1      &     85.2      &     0.70      &     54.9      &     86.0      &     0.66      &     57.9      &     81.6      &     0.66   \\ 
		&    BootEA*   &     67.8      &     91.2      &     0.76      &     68.2      &     89.8      &     0.76      &     64.8      &     91.9      &     0.74      &     66.5      &     87.1      &     0.73       \\ \midrule
		\multirow{3}{*}{GCN}      &   GCN-Align   &   43.1	&    71.3	&  0.53  & 31.3   &   57.5	& 0.40 &	37.3	&   70.9	&  0.49	& 32.1   & 	55.2 & 	0.40   \\ 
		&   VR-GCN &    39.1   &   73.1  	&  0.51  &	 28.5 	 &   59.3		&   0.39   &	35.3	&   69.6	&   0.47	& 	  31.5	& 	61.1 & 	  0.41    \\ 
		&     R-GCN     &   24.1   &  61.4   &  0.36    &    31.6  &   57.3      &  0.40     &   25.7   &   59.0     &   0.37    &   28.6     &   52.4   &   0.37       \\ \midrule
		\multirow{5}{*}{path}      &    PTransE   &  61.6    &   81.7  &   0.69     &   62.9    &   83.2     &       0.70  &       38.4        &      80.2       &       0.52       &       37.7       &       65.8        &      0.48     \\ 
		&   IPTransE*   &     43.5      &     74.5      &     0.54      &     23.6      &     51.3      &     0.33      &     42.9      &     78.3      &     0.55      &     34.0      &     63.2      &     0.44     \\ 
		&    Chains          &  64.9   &    88.3   &   0.73     &  71.3     &   91.8   &   0.79     &      70.6       &      92.6       &      0.78       &      70.7        &      86.4       &       0.76         \\ 
		&     RSN*      &     76.3      &     92.4      &     0.83      &      \textbf{82.6}       &       95.8       &      \textbf{0.87}       &     75.6      &     92.5      &     0.82      &     73.9      &     89.0      &     0.79     \\ \cmidrule{2-14}
		& \textbf{Interstellar} & \textbf{77.9} & \textbf{94.1} & \textbf{0.84} & {81.8} & \textbf{96.2} & \textbf{0.87 } & \textbf{77.3} & \textbf{94.7} & \textbf{0.84} & \textbf{75.3} & \textbf{90.3} & \textbf{0.81} \\ \bottomrule
	\end{tabular}
\end{table}

\subsection{Problem under parameter-sharing}
\label{app:evalprob}

\begin{figure}[H]
	\centering
	\subfigure[Entity alignment (DBP-WD).]
	{\includegraphics[width=0.45\textwidth]{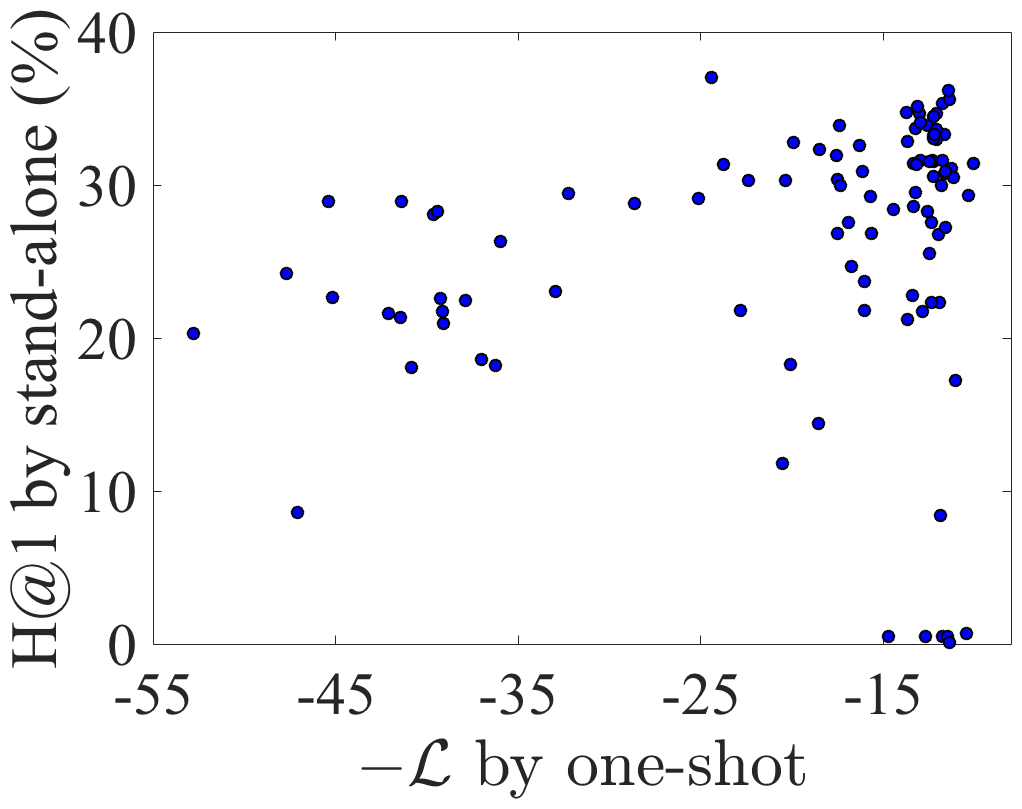}}
	\hfill
	\subfigure[Link prediction  (WN18-RR).]
	{\includegraphics[width=0.45\textwidth]{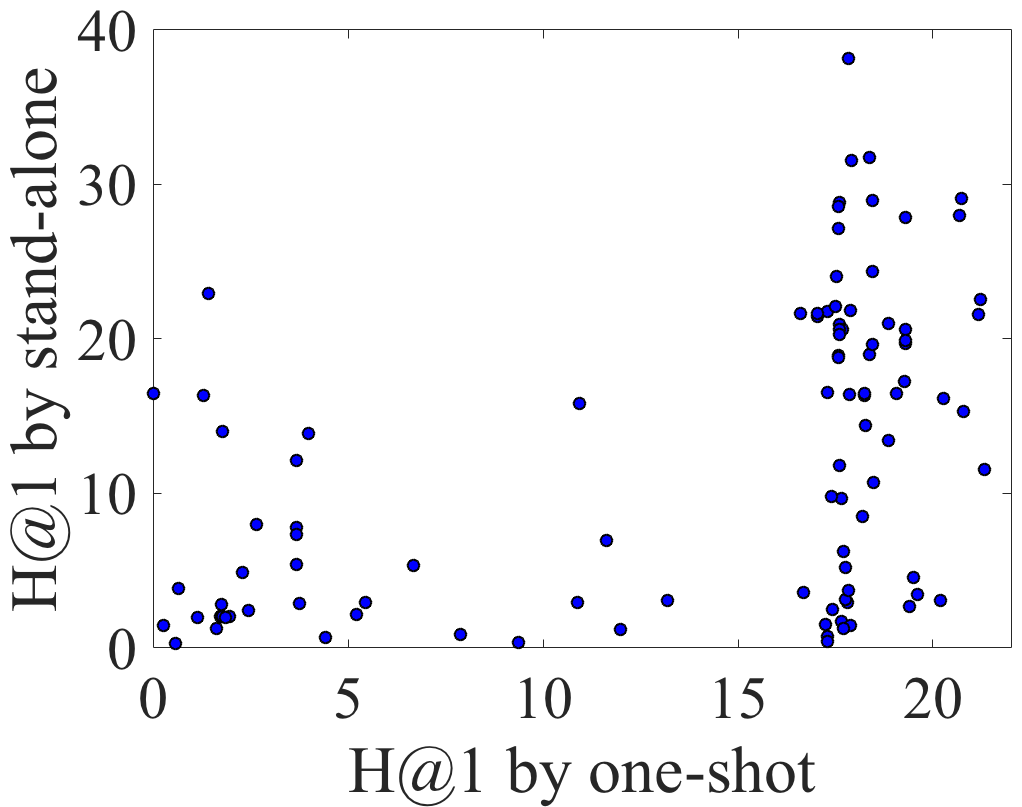}}
	\caption{Comparison of models trained with one-shot model and stand-alone model on the full search space.}
	\label{fig:ps:full}
\end{figure}

In this part, we empirically show the problem under parameter-sharing in the full search space $\mathcal A$
and how the micro-level space $\mathcal{A}_2$ solves this problem.
Following \cite{bender2018understanding,zela2019understanding},
we train the supernet in Figure~\ref{fig:struct} with parameters-sharing
and 1) randomly sample 100 architectures from $\mathcal A$ (one-shot);
2) re-train these sampled architectures from scratch (standard-alone).
If the top-performed architectures selected by one-shot approach also perform well by the stand-alone approach,
then parameter-sharing works here and vice the visa.
Based on the discussion in Appendix~\ref{ssec:ngimplement},
we use the loss on validation mini-batch as the measurement in the one-shot approach for entity alignment task.
And we use the metric, i.e. Hit@1, 
on validation mini-batch as the measurement in the stand-alone approach for link prediction task.
Figure~\ref{fig:ps:full} shows such a comparison in the full space $\mathcal A$.
As shown in the upper-right corner of each figure,
the top-performed architectures in one-shot approach
do not necessarily perform the best in stand-alone approach.
Therefore, searching architectures by one-shot approach in the full search space does not work
since parameter-sharing does not work here.

\begin{figure}[H]
	\centering
	
	\subfigure[Entity alignment (DBP-WD).]
	{\includegraphics[width=0.45\textwidth]{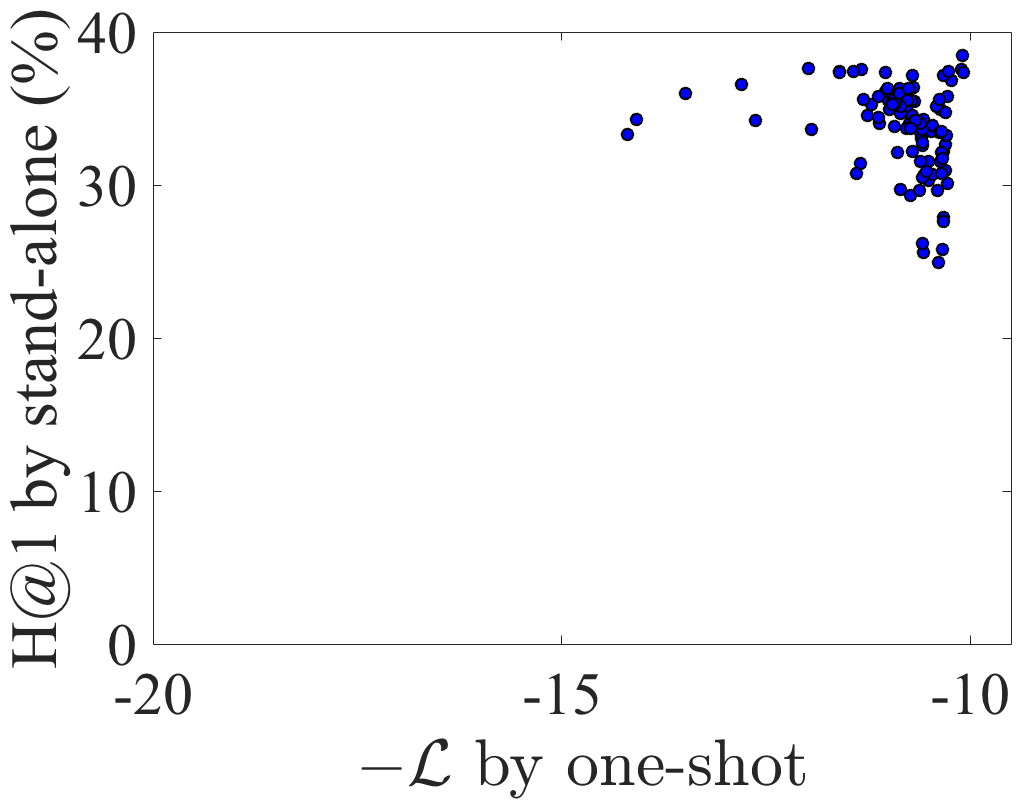}}
	\hfill
	\subfigure[Link prediction (WN18-RR).]
	{\includegraphics[width=0.45\textwidth]{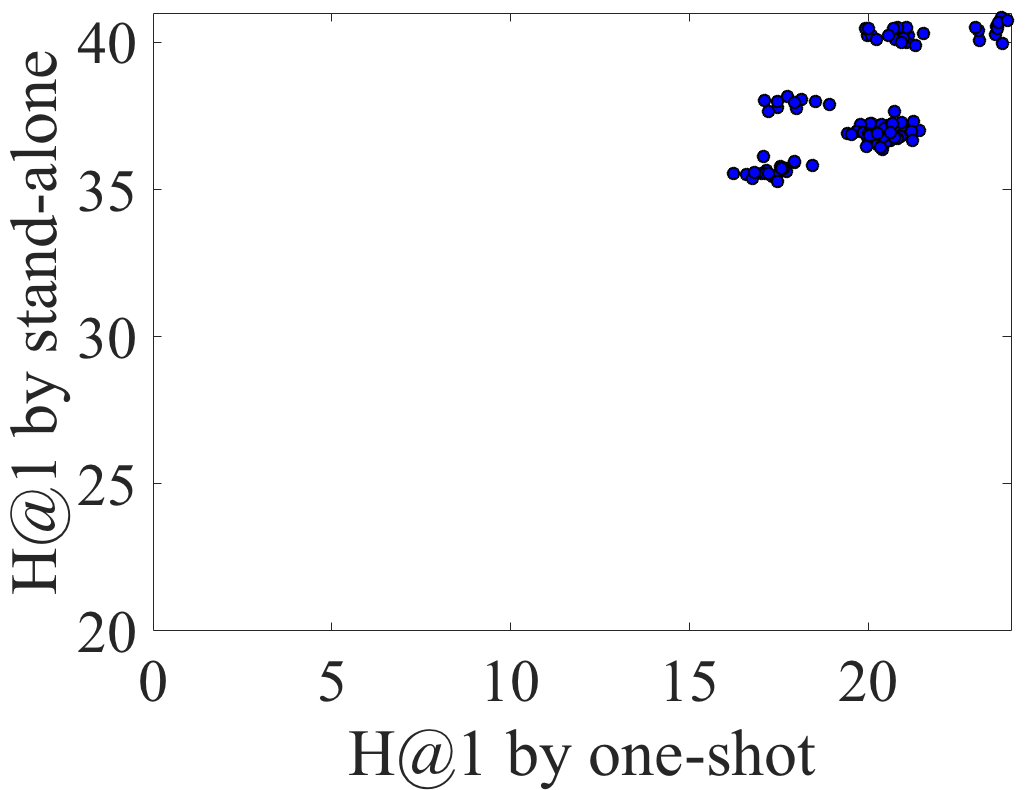}}
	\caption{Comparison of models trained with one-shot model and stand-alone model on the micro-level search space.}
	\label{fig:ps:micro}
\end{figure}

In order to take advantage of the efficiency in one-shot approach,
we split the search space into a macro-level space $\mathcal{A}_1$ and micro-level space $\mathcal{A}_2$.
To show how parameter-sharing works in the micro-level space,
we use the best macro-level architecture $\bm{\alpha}_1^*$ during macro-level searching,
and sample 100 architectures ${\bm \alpha}_2\in\mathcal{A}_2$ to get the architecture 
$\bm \alpha=[\bm{\alpha}_1^*, {\bm \alpha}_2]$.
As shown in Figure~\ref{fig:ps:micro},
the top performed architectures in one-shot approach also perform the best in stand-alone approach.
This observation verifies that 
1) parameter-sharing is more likely to work in a simpler search space as discussed in \cite{bender2018understanding,pham2018efficient,zela2019understanding};
2) the split of the macro and micro level space in our problem is reasonable;
3) searching architectures by the Hybrid-search algorithm takes the advantage of 
the stand-alone approach and the one-shot approach.

\subsection{Influence of path length}

\begin{figure}[H]
	\centering
	{\includegraphics[width=0.45\textwidth]{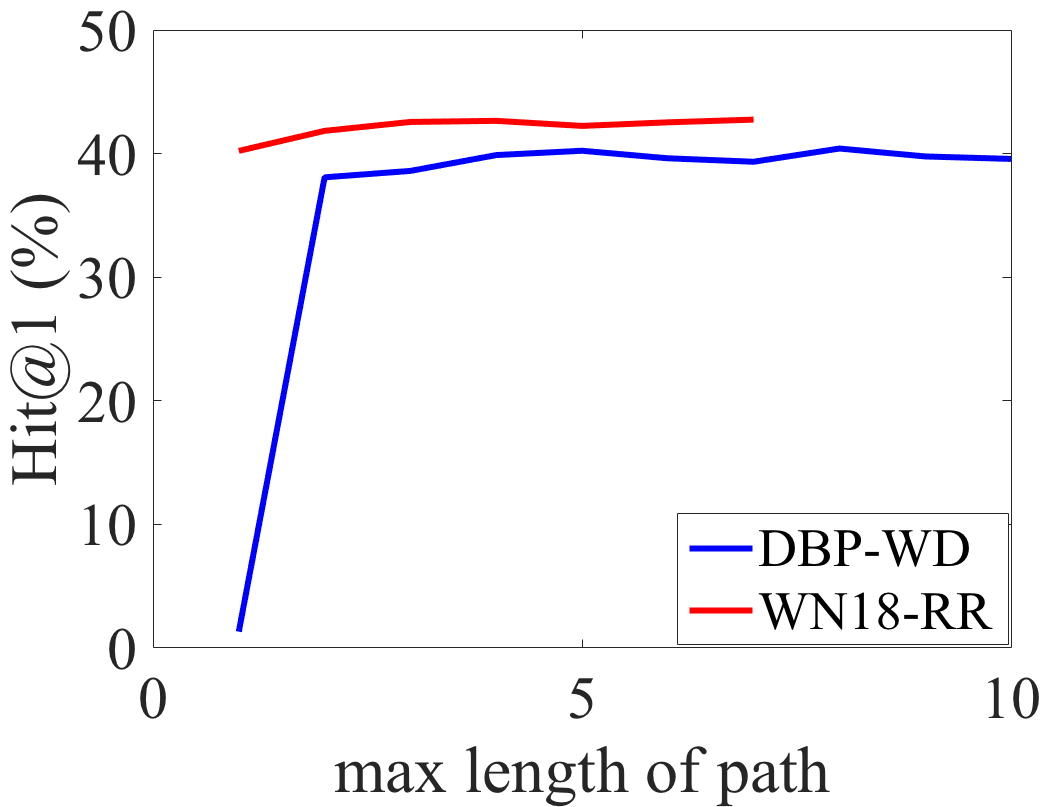}}
	\caption{The influence of path length.}
	\label{fig:length}
\end{figure}

In this part, we show the performance of the best architecture on each datasets with varied maximum length of the relational path
in Figure~\ref{fig:length}.
For entity alignment task, we vary the length from 1 to 10 for DBP-WD datasets (normal version);
for link prediction task, we vary the length from 1 to 7 for WN18-RR datasets.
We plot them in the same figure to make better comparison.
As shown,
when path length is $1$, the model performs very bad.
However, it has little influence on WN18-RR.
This verifies that long-term information is important for the entity alignment task
while short-term information is important for link prediction task.
For the other length, the performance is quite stable.

\section{Searched Models}
\label{app:models}

\subsection{Architectures in literature}
\label{app:literature}

\begin{figure}[ht]
	\centering
	\subfigure[TransE.]
	{\includegraphics[height = 2.6cm]{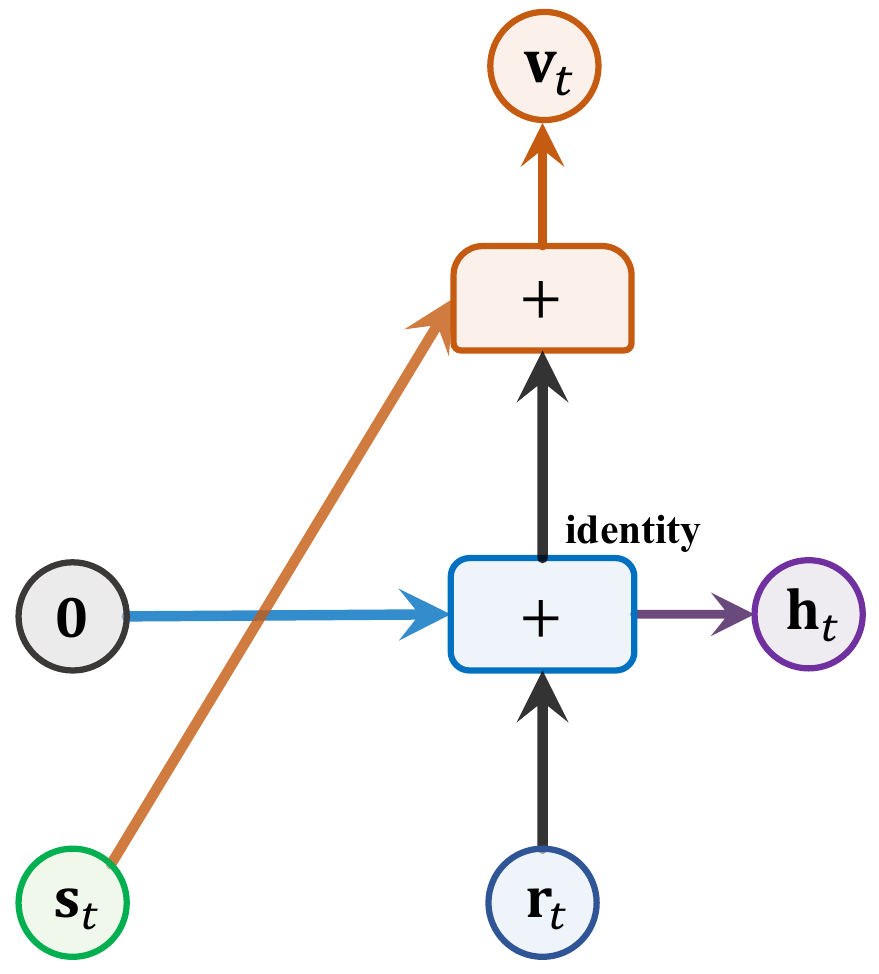}}
	\ \ 
	\subfigure[PTransE.]
	{\includegraphics[height = 2.6cm]{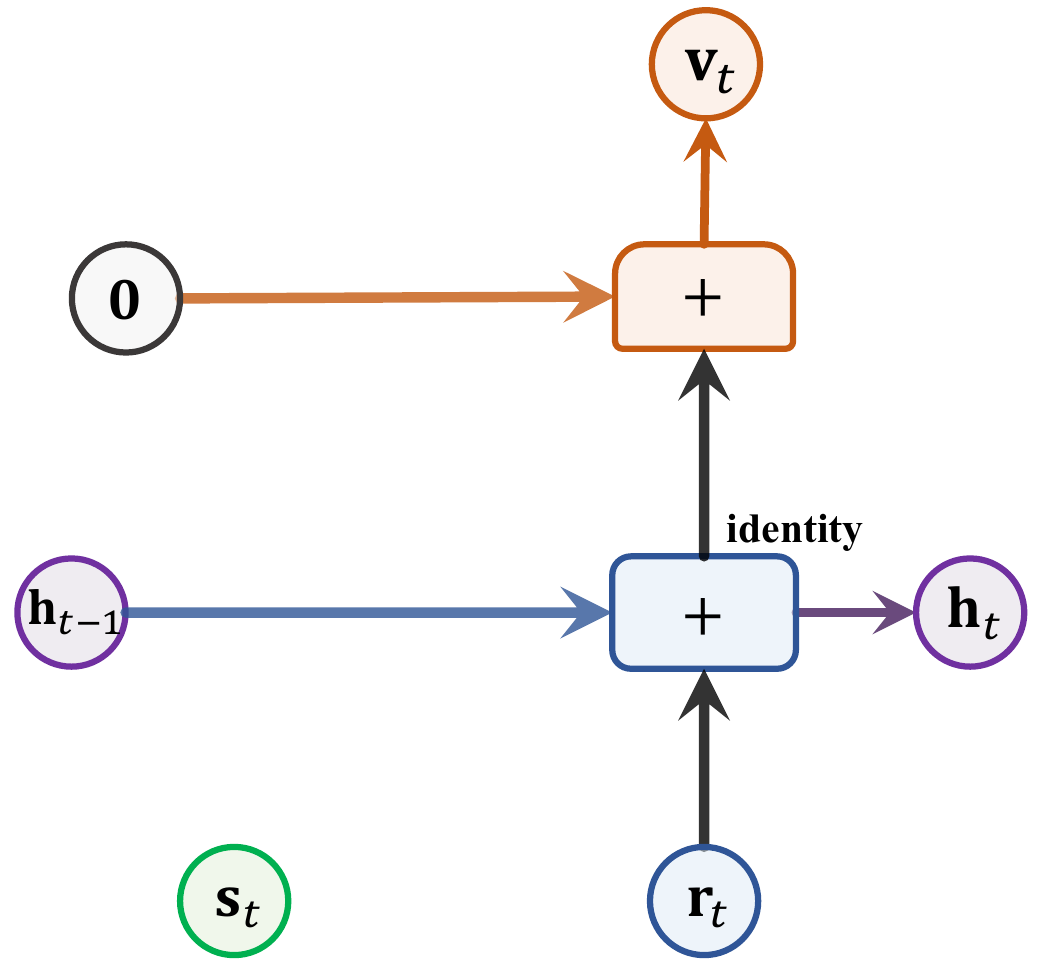}}
	\ \ 
	\subfigure[Chains.]
	{\includegraphics[height = 2.6cm]{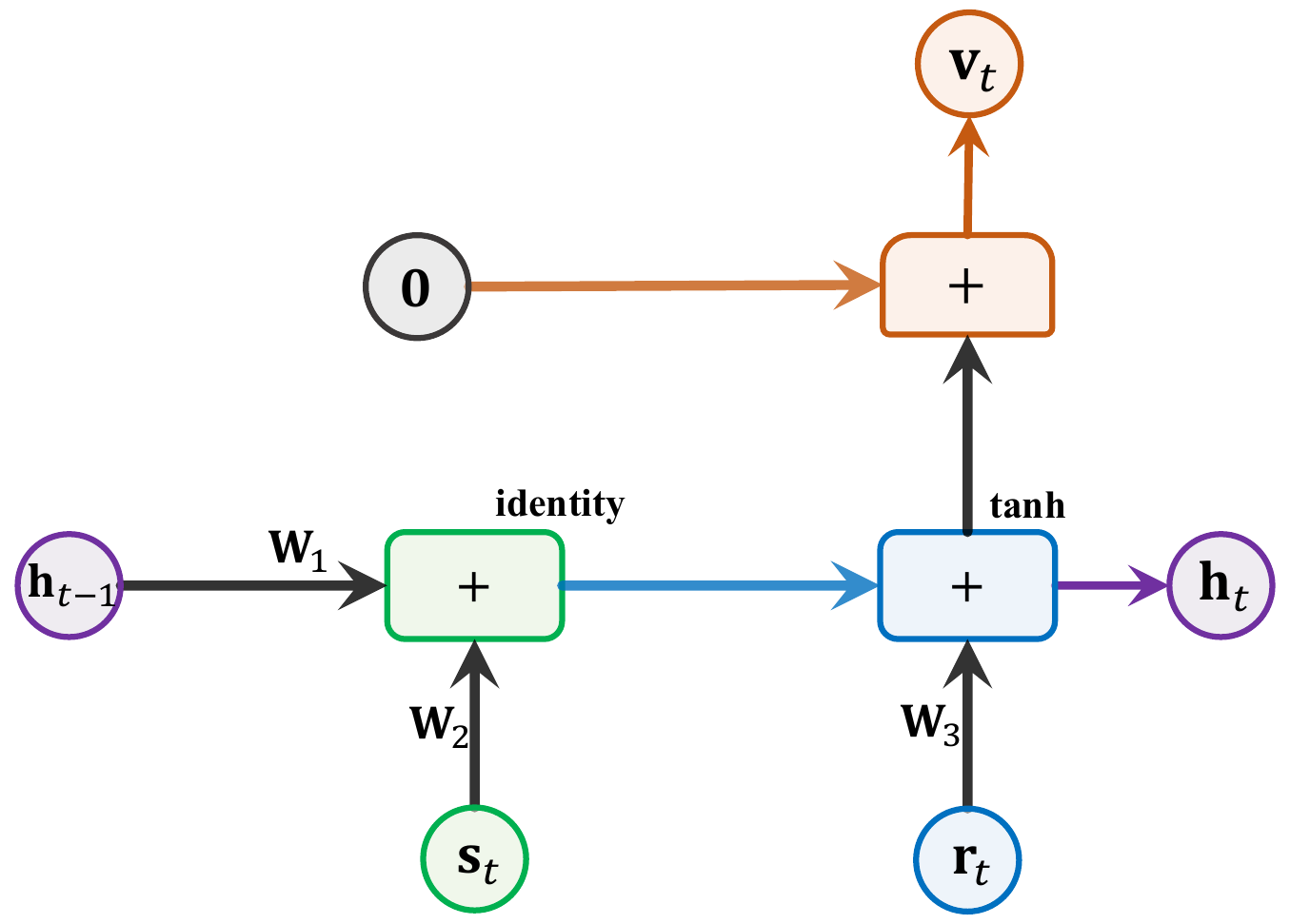}}
	\ \ 
	\subfigure[RSN.]
	{\includegraphics[height = 2.6cm]{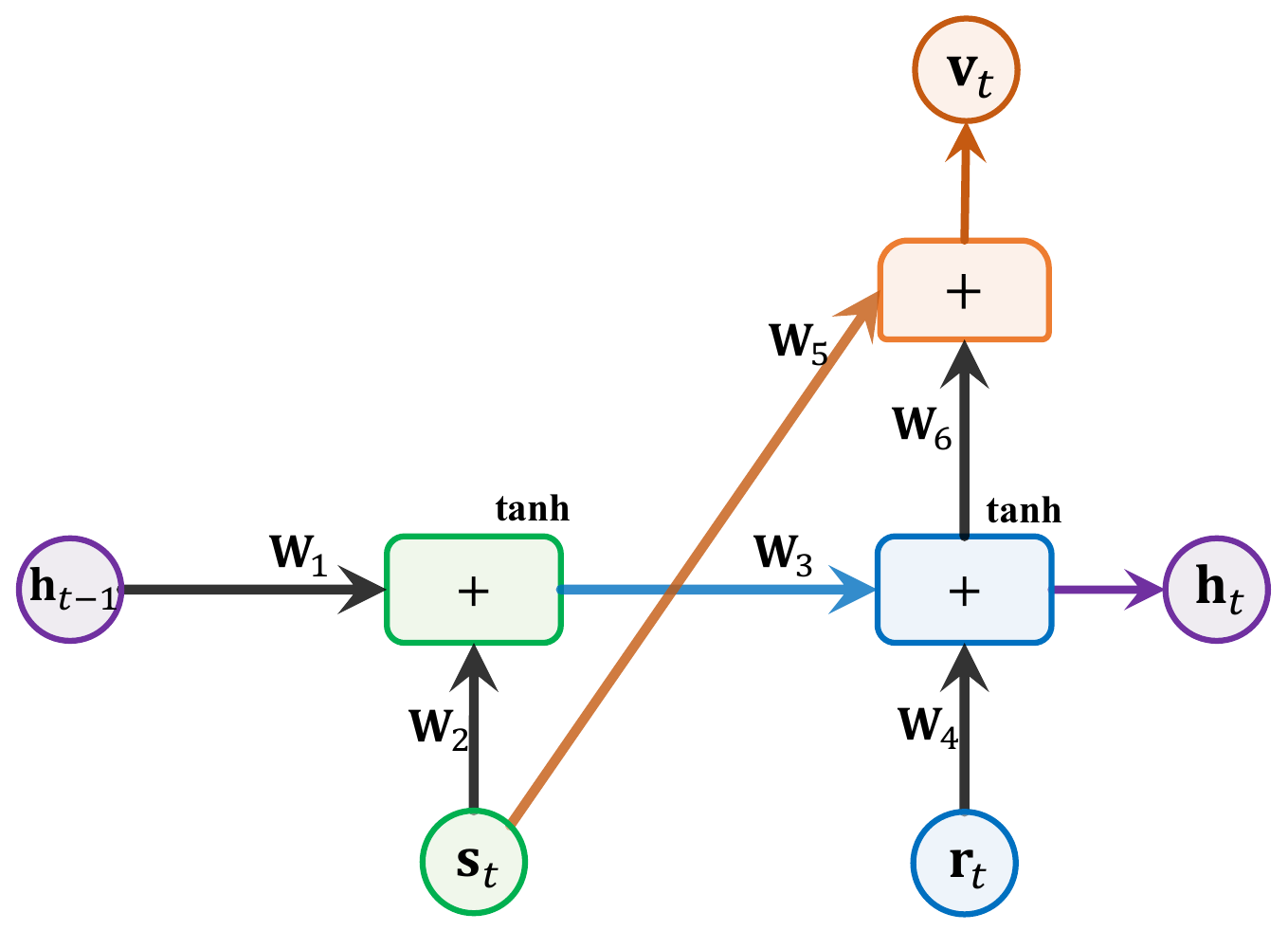}}
	\vspace{-3px}
	\caption{Graphical illustration of TransE, PTransE, Chains, RSN. 
		TransE is a triplet-based model thus it is non-recurrent. 
		PTransE recurrently processes the relational path without the intermediate entities.
		Chains simultaneously combines the entities and relations along the path,
		and 
		RSN designs a customized variant of RNN. 
		Each edge is an identity mapping, except an explicit mark of the weight matrix $\mathbf W_i$.
		Activation functions are shown in the upper right corner of the operators.}
	\label{fig:graphical}
\end{figure}

\subsection{Countries}
\label{app:countries}
\begin{figure}[H]
	\centering
	\subfigure[S1]
	{\includegraphics[height=3.0cm]{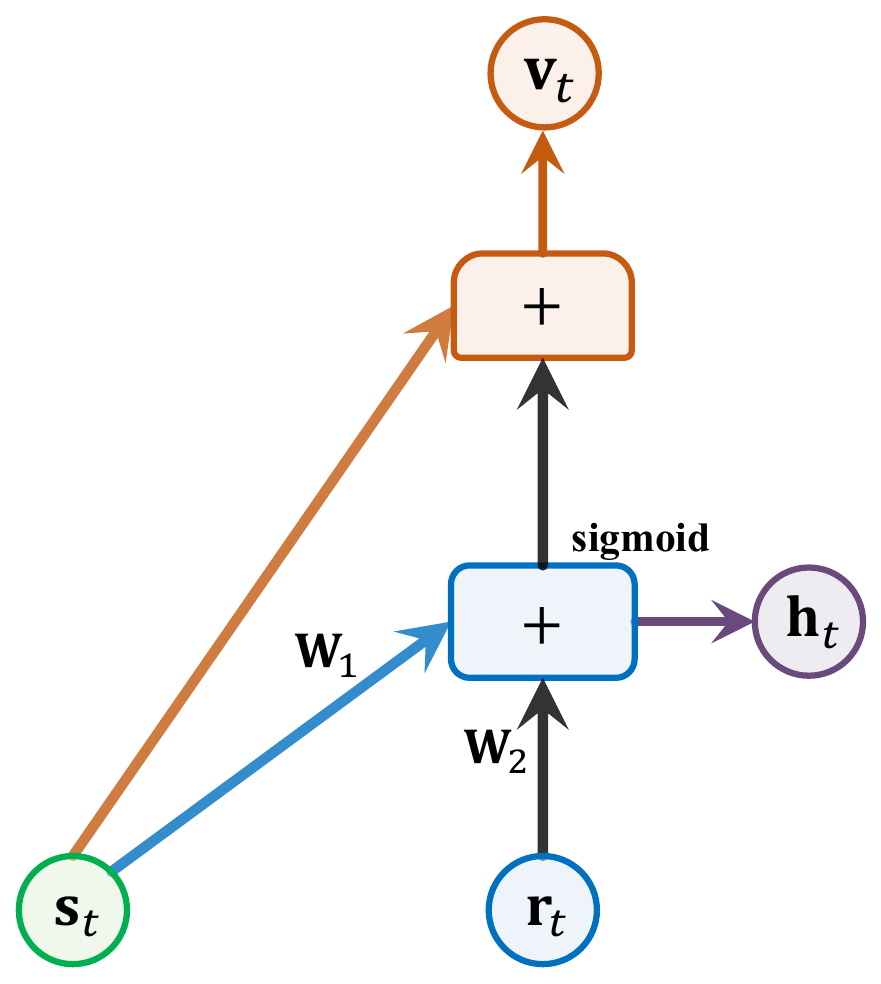}}
	\quad\quad
	\subfigure[S2]
	{\includegraphics[height=3.0cm]{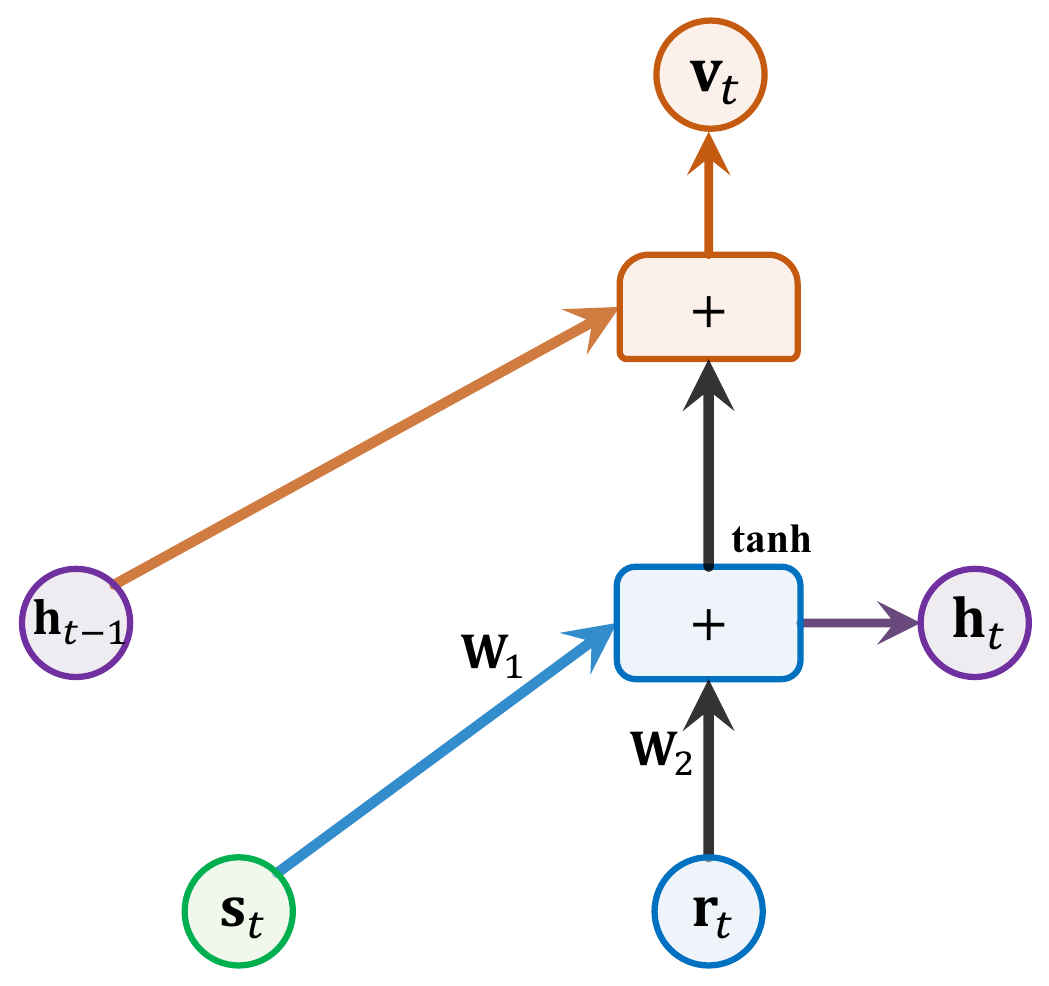}}
	\quad\quad
	\subfigure[S3]
	{\includegraphics[height=3.0cm]{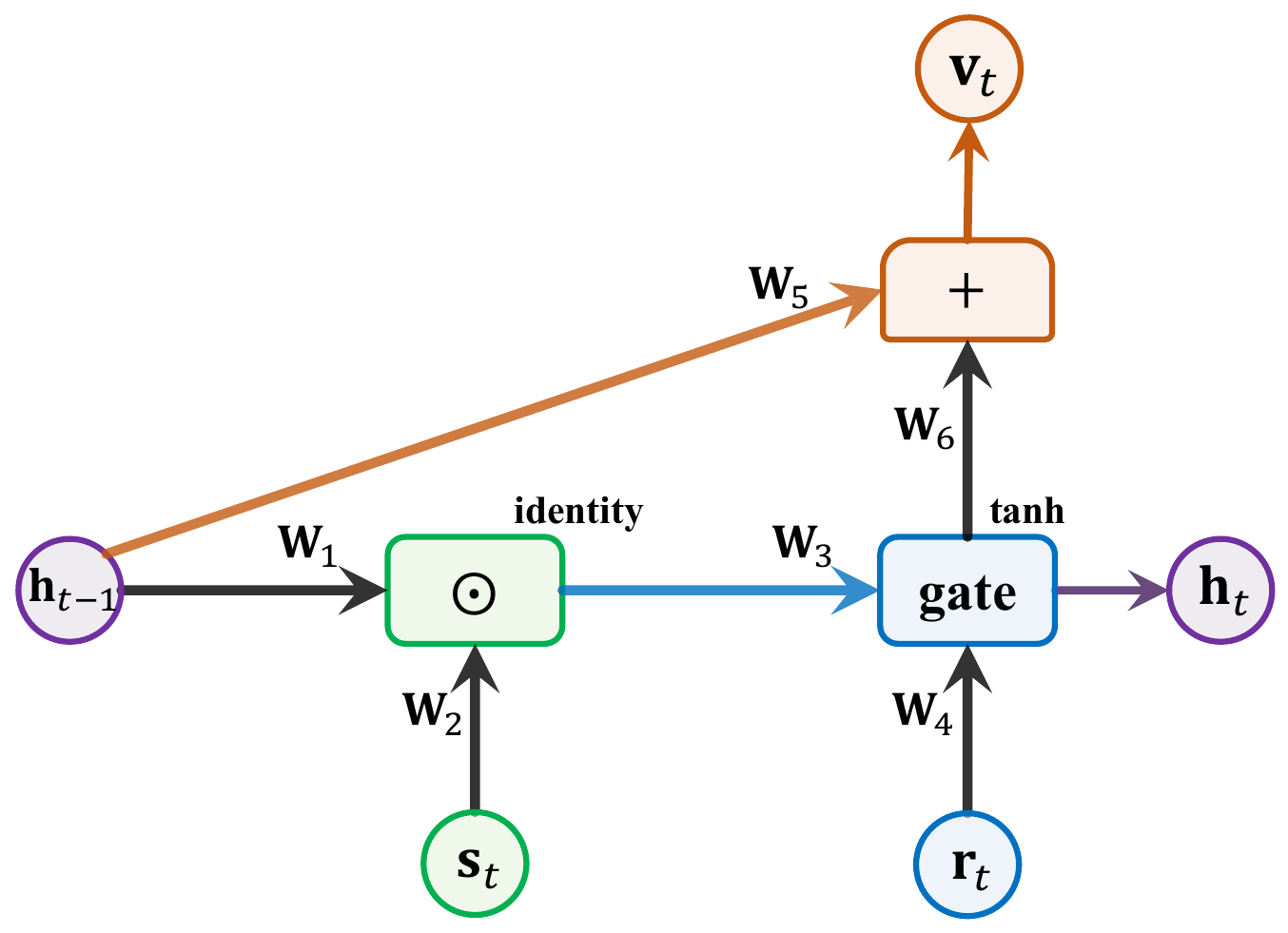}}
	\caption{Graphical representation of the searched $f$ on countries dataset.}
	\label{fig:country:best}
\end{figure}

We give the graphical illustration of architectures searched in Countries dataset in Figure~\ref{fig:country:best}.
The search procedure is conducted in the whole search space 
rather than the four patterns P1-P4 in Figure~\ref{fig:countries}.
We can see that in Figure~\ref{fig:country:best}, (a) belongs to P1, (b) belongs to P2
and (c) belongs to P4.
These results further verify that
Interstellar can adaptively search architectures for specific KG tasks and datasets.

\subsection{Entity Alignment}
\label{app:EAmodels}
\begin{figure}[ht]
	\centering
	\subfigure[DBP-WD]
	{\includegraphics[height = 2.5cm]{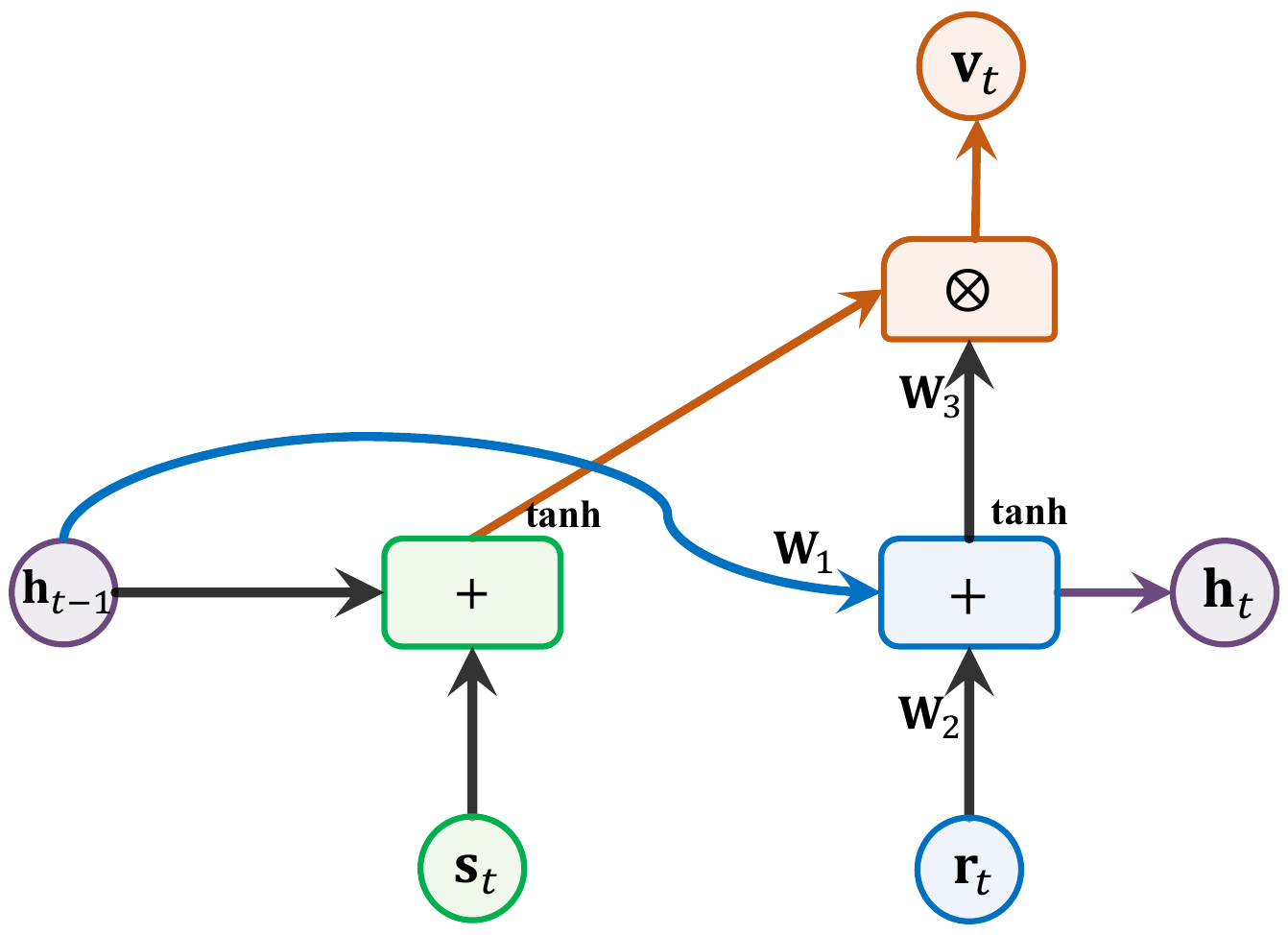}}
	\ 
	\subfigure[DBP-YG]
	{\includegraphics[height = 2.5cm]{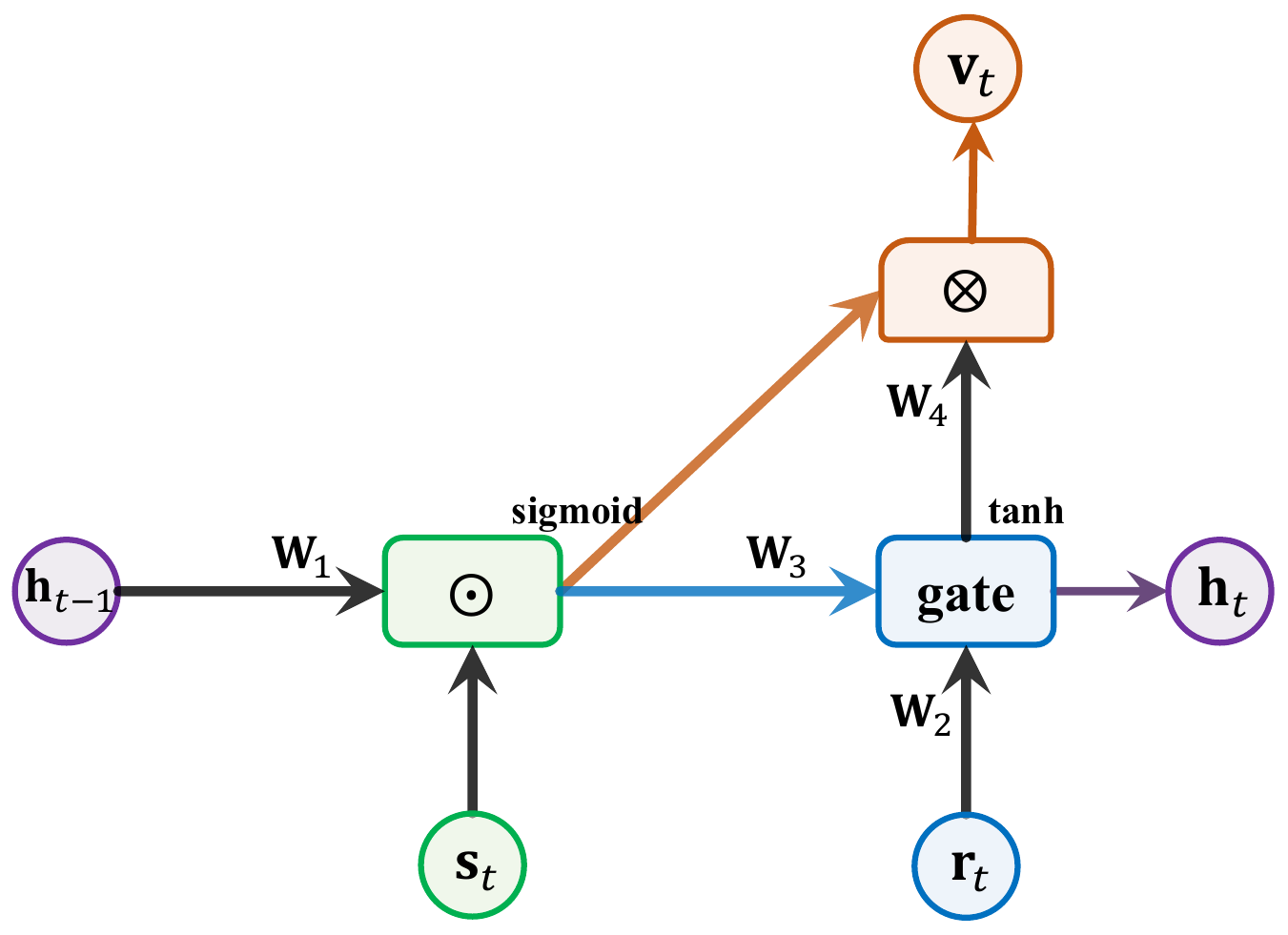}}
	\subfigure[EN-FR]
	{\includegraphics[height = 2.5cm]{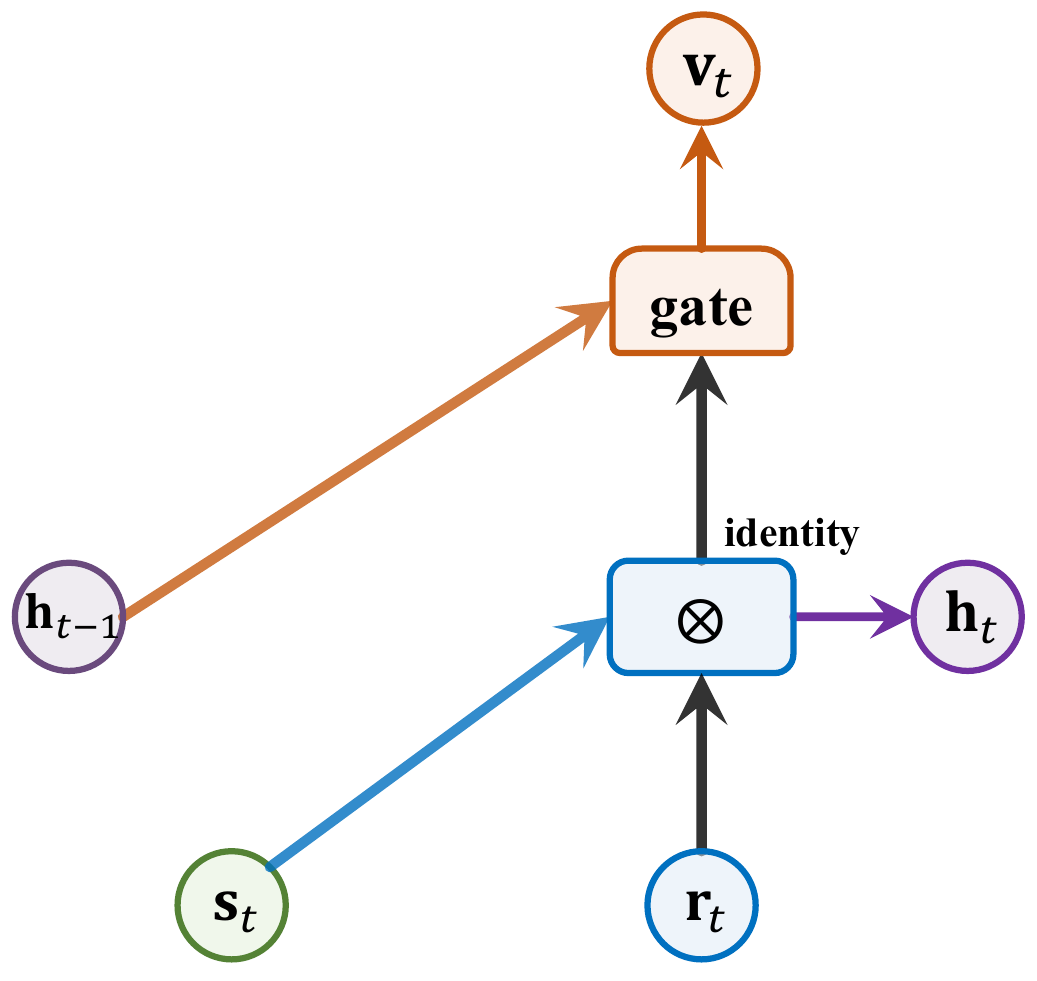}}
	\quad
	\subfigure[EN-DE]
	{\includegraphics[height = 2.5cm]{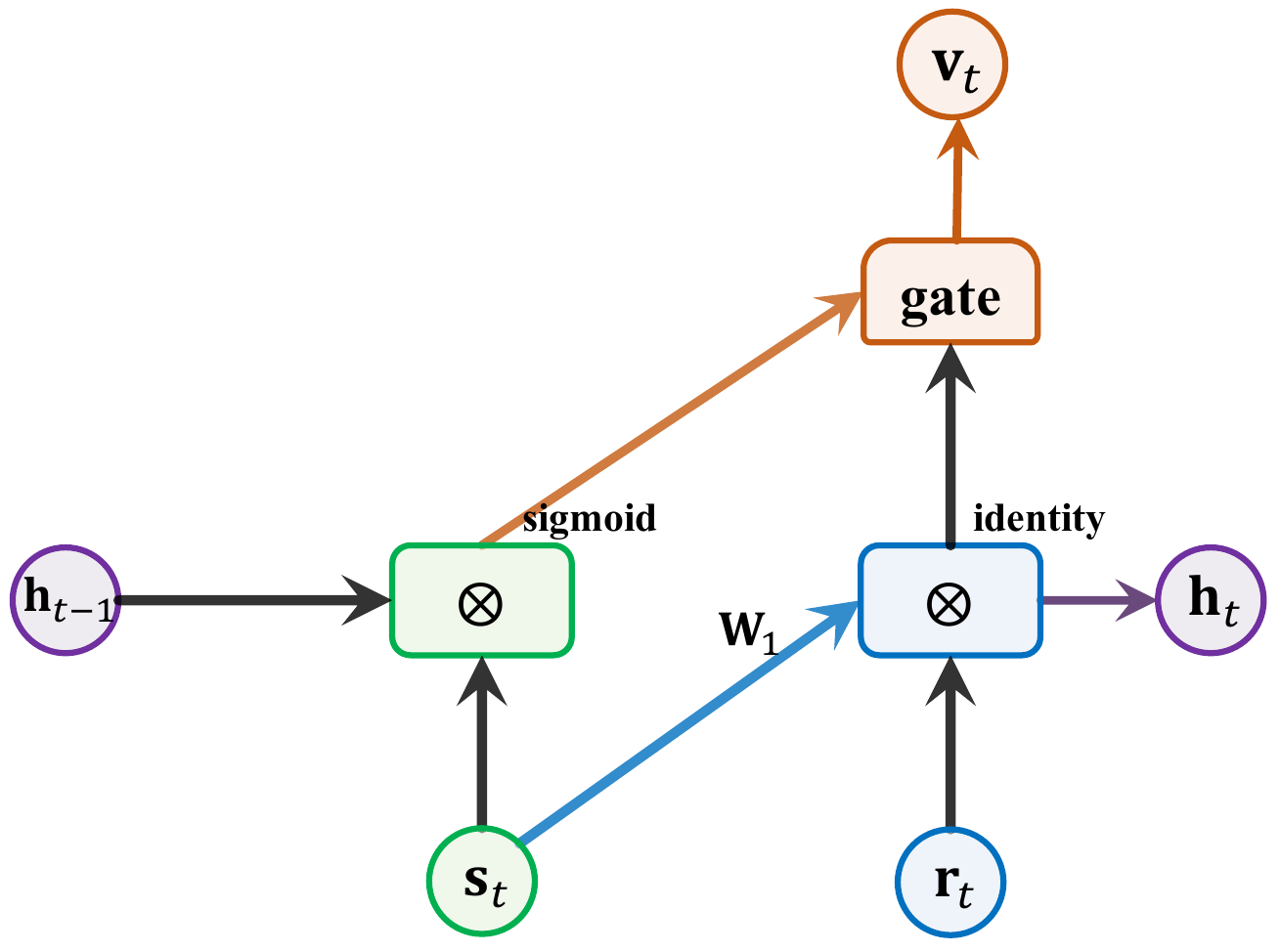}}
	\caption{Graphical representation of the searched recurrent network $f$ on each datasets in entity alignment task (Normal version).}
	\label{fig:align}
\end{figure}

The searched models by Interstellar,
which consistently perform the best on each dataset,
are graphically illustrated in Figure~\ref{fig:align}.
As shown,
all of 
the searched recurrent networks 
processing recurrent information in $\mathbf h_{t-1}$,
subject entity embedding $\mathbf s_t$ 
and relation embedding $\mathbf r_t$ together.
They have different connections, different composition operators and different activation functions,
even though the searching starts with the same random seed \textit{1234}.

More interestingly,
the searched architectures in DBP-WD and DBP-YG can model multiple triplets simultaneously.
While the architectures in EN-FR and EN-DE models two adjacent triplets each time.
The reason is that, 
in the first two datasets, the KGs are from different sources, thus the distribution is various \cite{guo2019learning}.
In comparison, EN-FR and EN-DE are two cross-lingual datasets,
which should have more similar distributions in each KG.
Therefore, we need to model the longer term information on DBP-WD and DBP-YG.
In comparison, modeling the shorter term information like two triplets together is better for EN-FR and EN-DE.
In this way, the model can focus more on learning short-term semantic in a range of two triplets.


\begin{figure}[ht]
	\centering
	\subfigure[DBP-WD]
	{\includegraphics[height = 2.5cm]{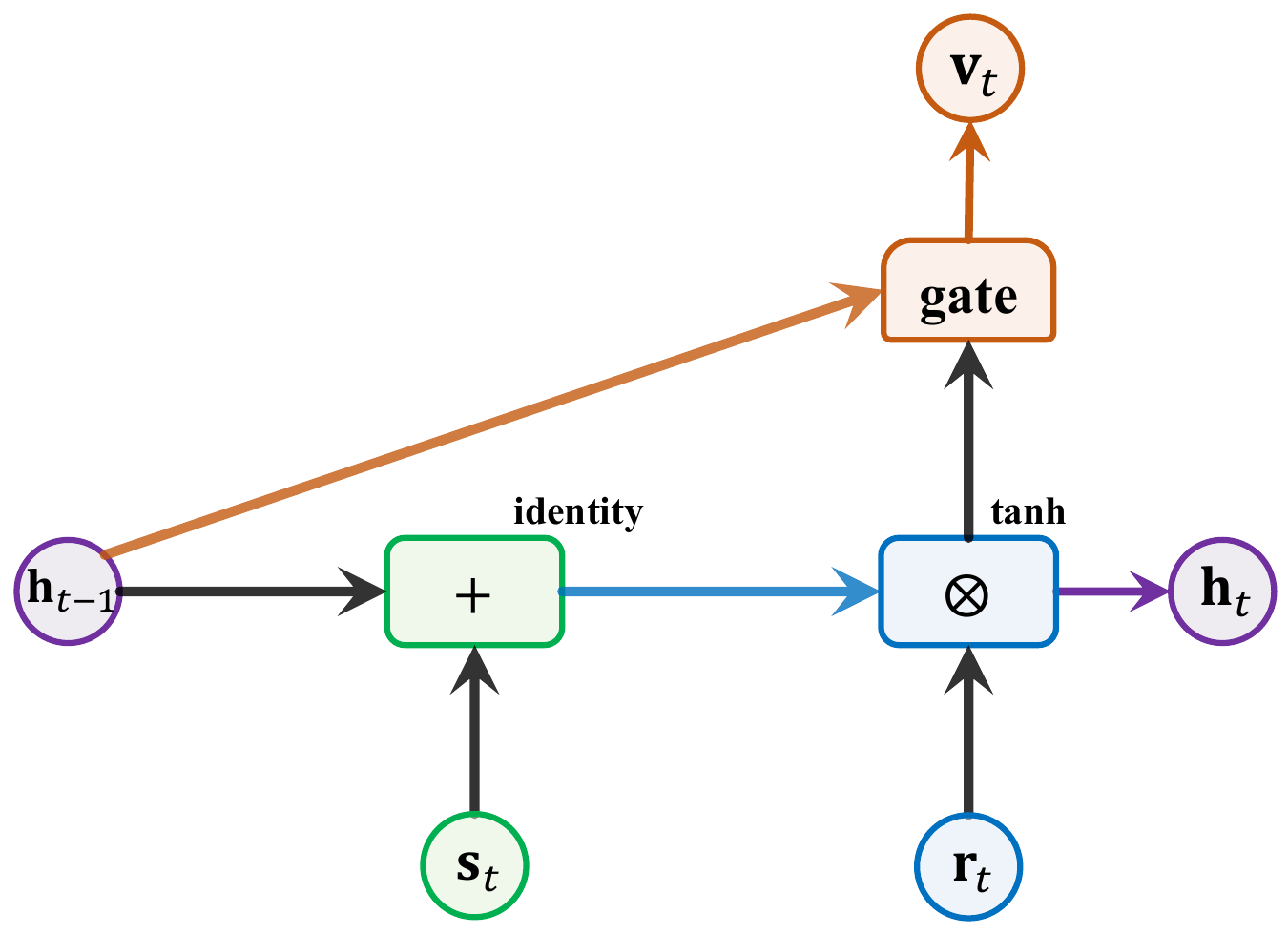}}
	\ 
	\subfigure[DBP-YG]
	{\includegraphics[height = 2.5cm]{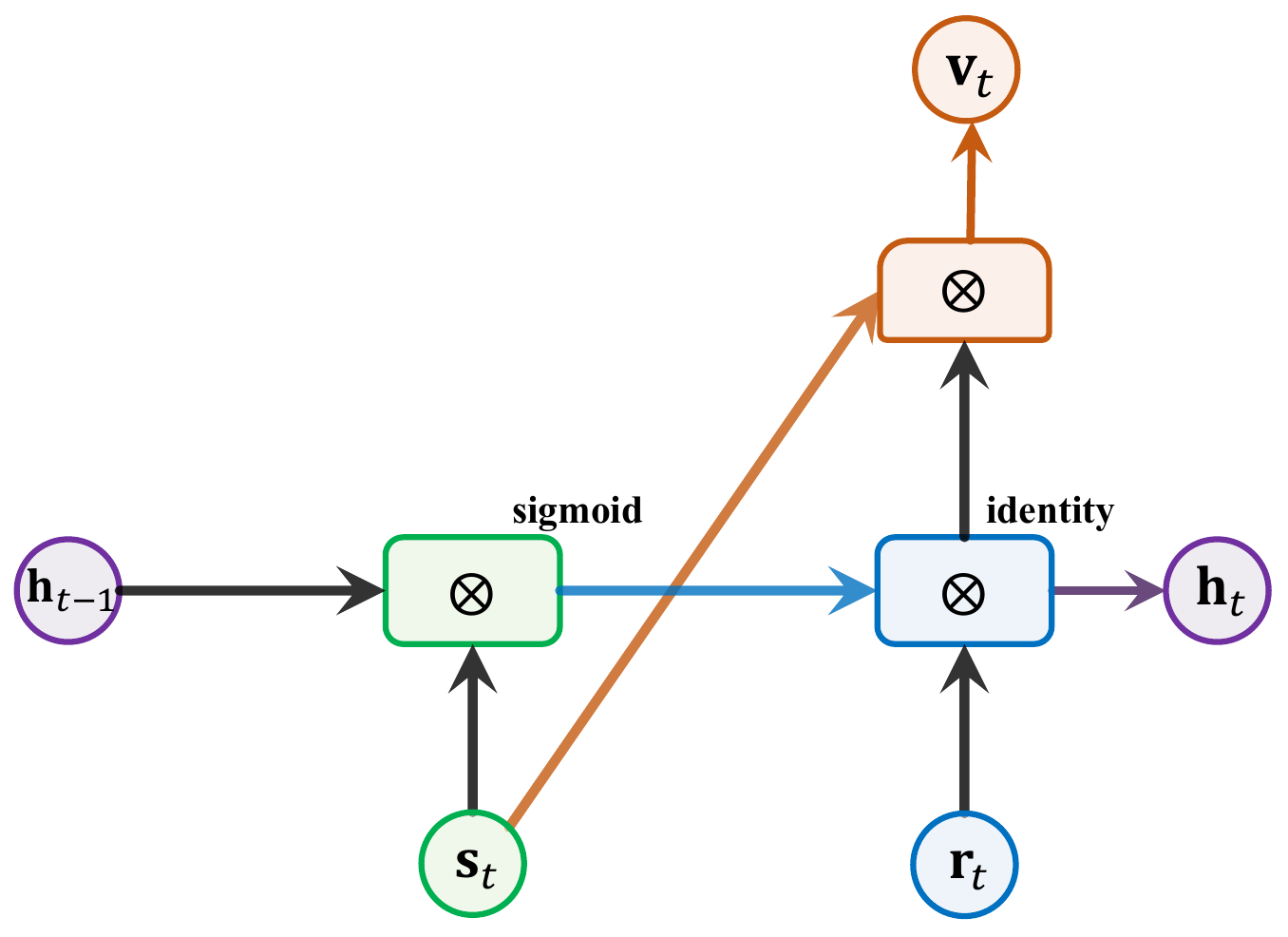}}
		\quad
	\subfigure[EN-FR]
	{\includegraphics[height = 2.5cm]{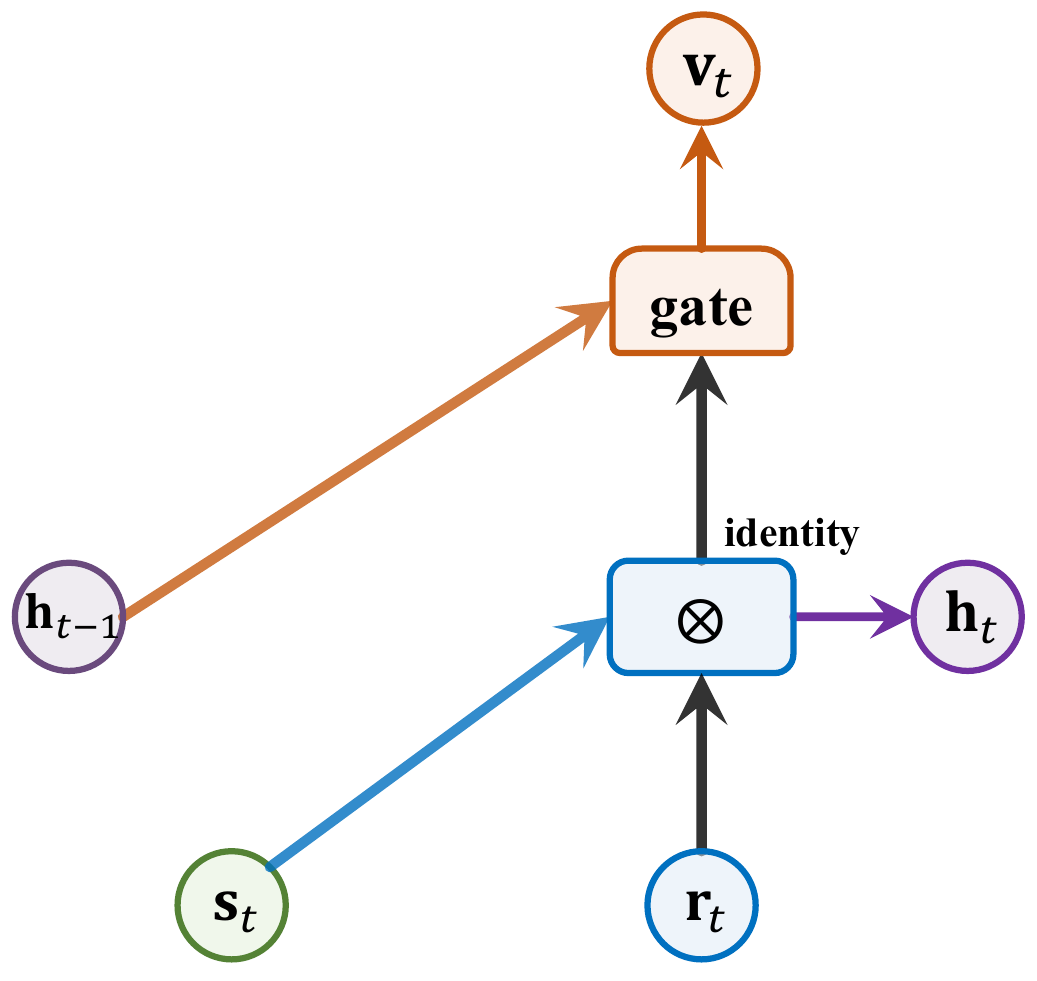}}
	\quad
	\subfigure[EN-DE]
	{\includegraphics[height = 2.5cm]{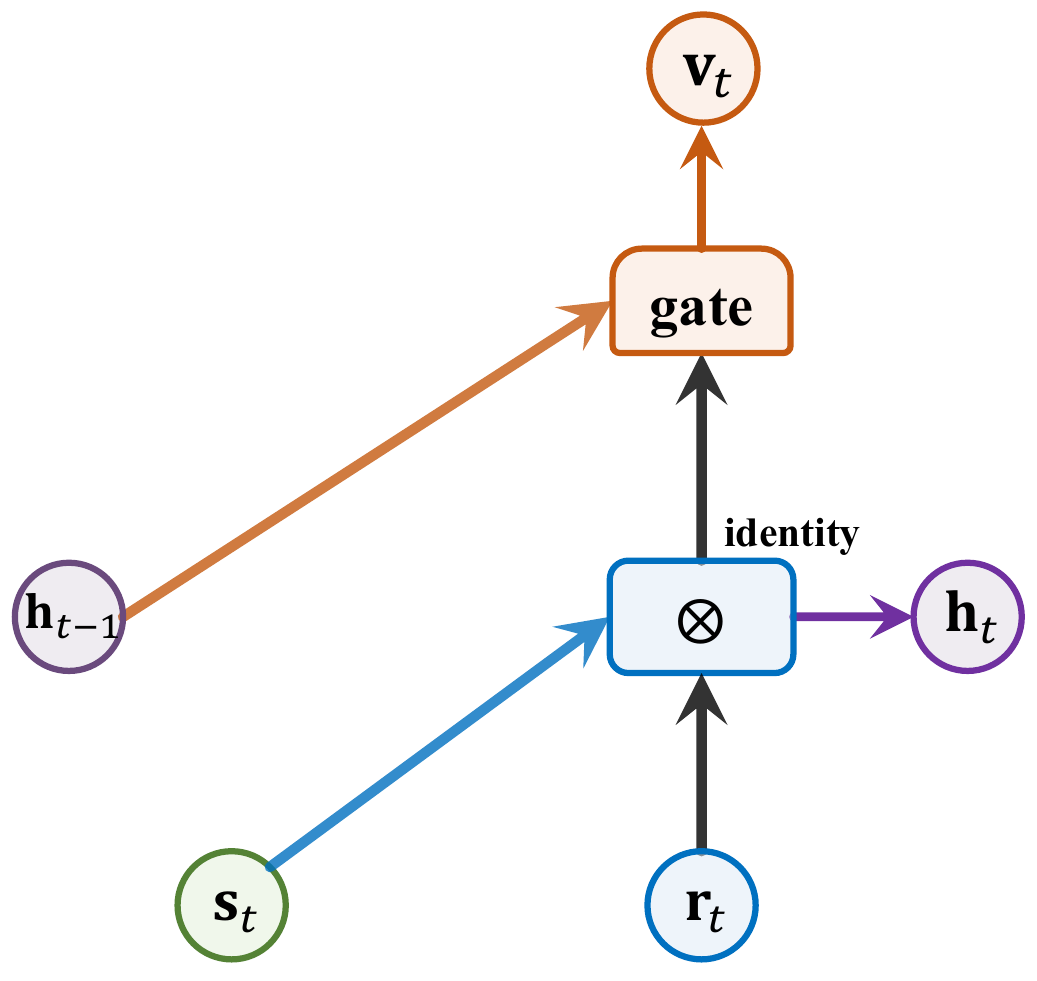}}
	\caption{Graphical representation of the searched recurrent network $f$ on each datasets in entity alignment task (Dense version).}
	\label{fig:align2}
\end{figure}

\subsection{Link Prediction}
\label{app:LPmodels}

\begin{figure}[H]
	\centering
	\subfigure[WN18-RR]
	{\includegraphics[height=3.2cm]{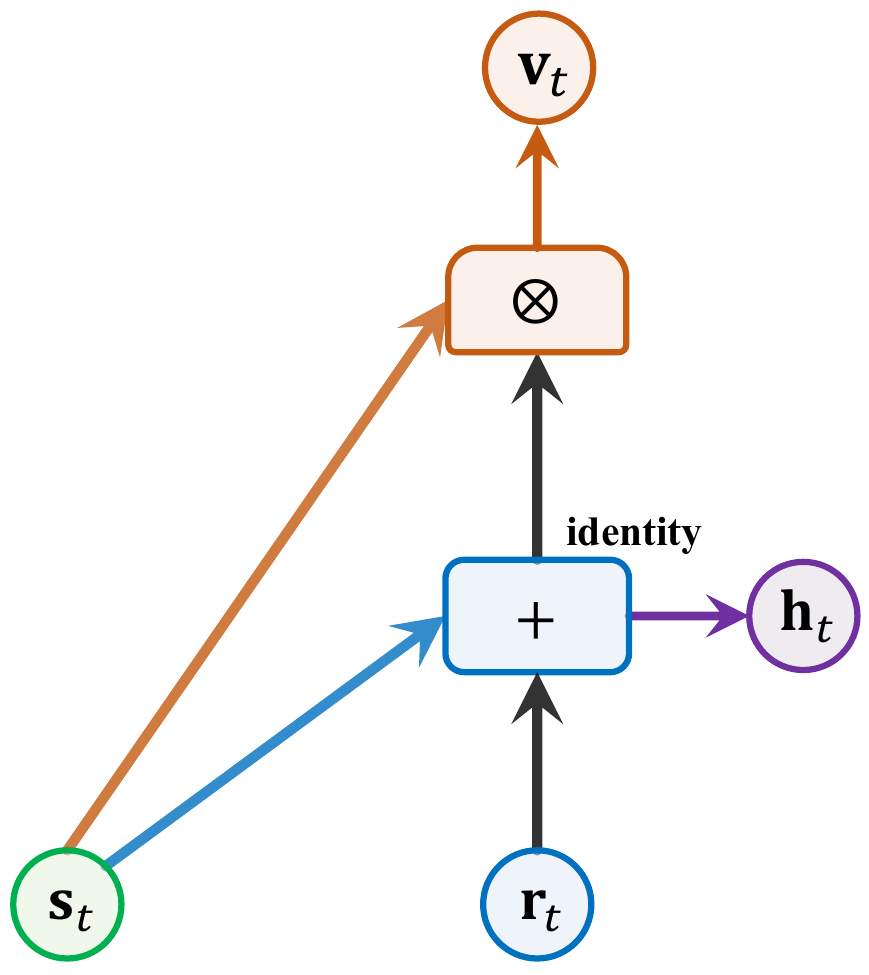}}
	\quad \quad
	\subfigure[FB15k-237]
	{\includegraphics[height=3.2cm]{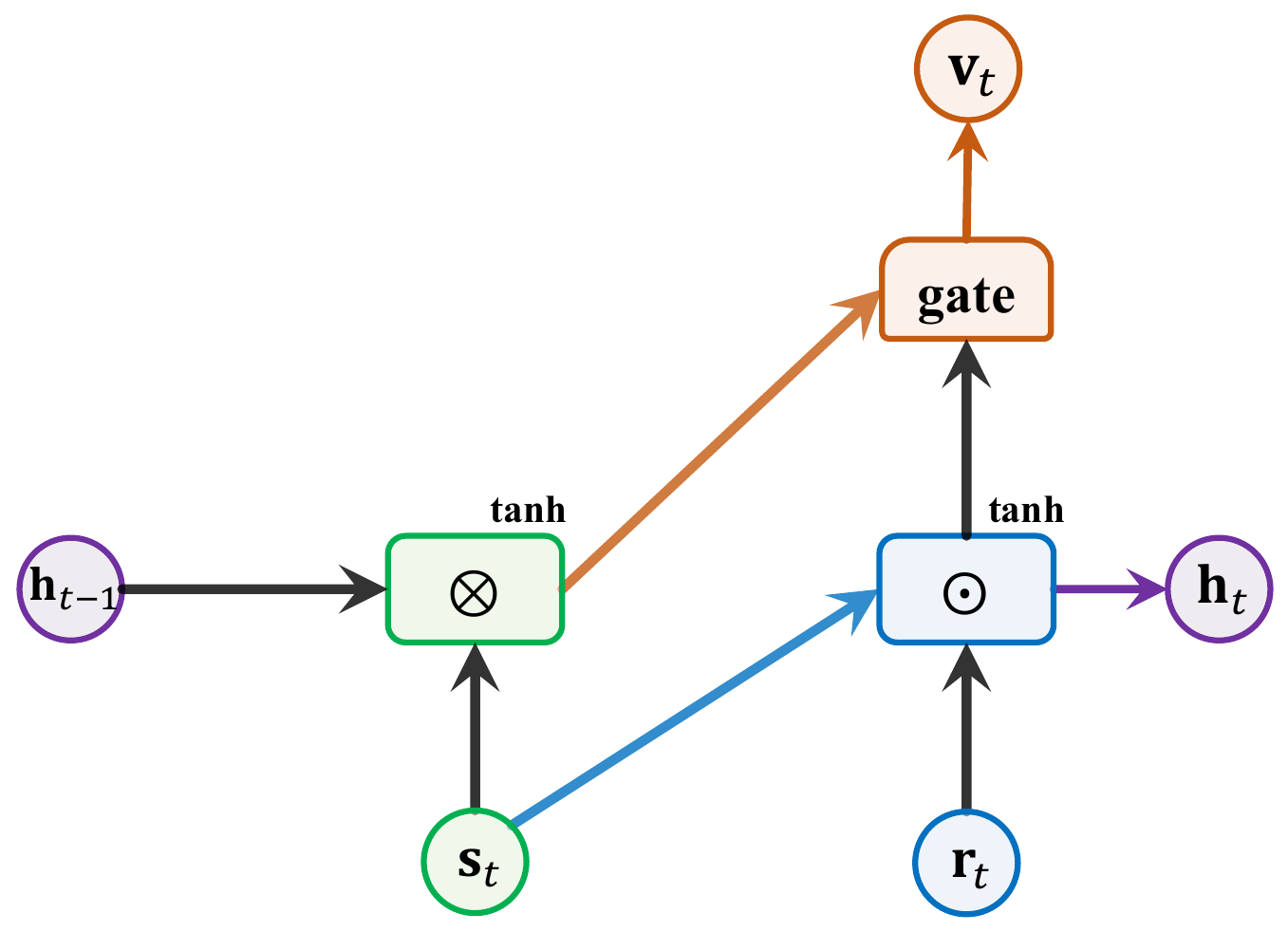}}
	\caption{Graphical representation of the searched $f$ on each datasets in link prediction task.}
	\label{fig:pred}
\end{figure}

The best architectures searched in link prediction tasks are given in Figure~\ref{fig:pred}.
In order to illustrate the models we searched,
we make a statistics of the distance, 
i.e. the shortest path distance when regarding the KG as a simple undirected graph,
between two entities in the validation and testing set in Table~\ref{tab:nhop}.
As shown, most of the triplets have distance less than 3.
Besides,
as indicated by the performance on the two datasets,
we infer that triplets far away from 3-hop are very challenging to model.
At least in this task,
triplets less or equal than 3 hops are the main focus for different models.
This also explains why RSN,
which processes long relational path,
does not perform well in the link prediction task.
The searched models in Figure~\ref{fig:pred} do not directly consider long-term structures.
Instead,
the architecture on WN18-RR models one triplet,
and the architecture on FB15k-237 focuses on modeling two consecutive triplets.
These are consistent with the statistics in Table~\ref{tab:nhop},
where WN18-RR has more one-hop and FB15k-237 has more two-hop triplets.

\begin{table}[H]
	\centering
	\caption{Percentage of the $n$-hop triplets in validation and testing datasets.}
	\label{tab:nhop}
	\begin{tabular}{c|c|c|c|c|c}
		\toprule
		\multicolumn{2}{c|}{\multirow{2}{*}{Datasets}} & \multicolumn{4}{c}{Hops}                              \\ \cmidrule{3-6} 
		\multicolumn{2}{c|}{}                          & $\leq 1$ & 2      & 3      & $\geq 4$      \\ \midrule
		\multirow{2}{*}{WN18-RR}       & validation    & 35.5\%       & 8.8\%  & 22.2\% & 33.5\%                \\ 
		& testing       & 35.0\%       & 9.3\%  & 21.4\% & 34.3\%                \\ \midrule
		\multirow{2}{*}{FB15k-237}     & validation    & 0\%          & 73.2\% & 26.1\% & 0.7\%                 \\ 
		& testing       &  0\%            &  73.4\%      &  26.8\%      & 0.8\%   \\ \bottomrule
	\end{tabular}
\end{table}

%
%

\end{document}